\documentclass[twoside,11pt]{article}
\usepackage[utf8]{inputenc}
\usepackage[T1]{fontenc}
\usepackage{amsmath,amssymb,amsthm}
\usepackage{mathabx,mathrsfs}
\usepackage{jair, rawfonts}
\usepackage{apalike}

\usepackage[authoryear,round,longnamesfirst]{natbib}
\ShortHeadings{Leveraging Pre-trained Models for Failure Analysis Triplets Generation}
{Kenneth, Anis, Mireille \& Xavier}
\firstpageno{1}
\usepackage{makecell}
\usepackage[hashEnumerators,smartEllipses]{markdown}
\usepackage{subfig}
\usepackage{caption}
\usepackage{mathabx}
\usepackage{wrapfig}
\usepackage{tikz}
\usepackage{animate}
\usepackage{graphicx}
\usepackage{url}
\usepackage{booktabs}
\usepackage{url}
\usepackage{booktabs}
\usepackage{mathtools, xparse}
\usepackage{epstopdf}
\usepackage{float}
\usepackage{multirow}
\usepackage{pdflscape}
\usepackage{adjustbox}
\usepackage{bm}
\usepackage{svg}
\usepackage{relsize}
\usepackage{algcompatible}
\usepackage{colortbl}
\usepackage{xcolor,colortbl, pifont}
\usepackage{stfloats}
\usepackage{listings}
\definecolor{codegreen}{rgb}{0,0.6,0}
\definecolor{codegray}{rgb}{0.5,0.5,0.5}
\definecolor{codepurple}{rgb}{0.58,0,0.82}
\definecolor{backcolour}{rgb}{0.95,0.95,0.95}
\definecolor{forestgreen}{RGB}{34,139,34}
\definecolor{orangered}{RGB}{239,134,64}
\definecolor{darkblue}{rgb}{0.0,0.0,0.6}
\definecolor{gray}{rgb}{0.4,0.4,0.4}

\lstdefinestyle{mystyle}{
    backgroundcolor=\color{backcolour},   
    commentstyle=\color{codegreen},
    keywordstyle=\color{darkblue},
    numberstyle=\tiny\color{codegray},
    stringstyle=\color{codepurple},
    basicstyle=\ttfamily\tiny,
    breakatwhitespace=false,         
    breaklines=true,                 
    captionpos=none,                    
    keepspaces=true,  
    otherkeywords={self},
    numbersep=5pt,                  
    showspaces=false,                
    showstringspaces=false,
    showtabs=false,                  
    tabsize=2,
}

\newcommand{\comment}[1]{}
\usetikzlibrary{matrix,shapes,arrows,fit}
\begin{document}

\title{Leveraging Pre-trained Models for Failure Analysis Triplets Generation}

\author{\name Kenneth Ezukwoke$^{\dagger}$ \email ifeanyi.ezukwoke@emse.fr\\
        \name Anis Hoayek$^{\dagger}$ \email anis.hoayek@emse.fr\\
        \name Mireille Batton-Hubert$^{\dagger}$ \email batton@emse.fr\\
        \name Xavier Boucher$^{\ddagger}$ \email boucher@emse.fr \\
        \addr Department of Mathematics and Industrial  Engineering$^{\dagger}$\\
        \addr    Center for Biomedical and Healthcare Engineering$^{\ddagger}$\\
             Mines Saint-Etienne\\
             Univ. Clermont Auvergne, CNRS UMR 6158 LIMOS\\
             Saint-Etienne, France\\
       \AND
       \name Pascal Gounet \email pascal.gounet@st.com \\
       \name Jerome Adrian \email jerome.adrian@st.com \\
       \addr Failure Analysis Laboratory\\
             STMicroelectronics\\
             Grenoble, France
             }


\maketitle

\begin{abstract}
Pre-trained Language Models recently gained traction in the Natural Language Processing (NLP) domain for text summarization, generation and question answering tasks. This stems from the innovation introduced in Transformer models and their overwhelming performance compared with Recurrent Neural Network Models (Long Short Term Memory (LSTM)). In this paper, we leverage the attention mechanism of pre-trained causal language models such as Transformer model for the downstream task of generating Failure Analysis Triplets (FATs) - a sequence of steps for analyzing defected components in the semiconductor industry. We compare different transformer model for this generative task and observe that Generative Pre-trained Transformer 2 (GPT2) outperformed other transformer model for the failure analysis triplet generation (FATG) task. In particular, we observe that GPT2 (trained on 1.5B parameters) outperforms pre-trained BERT, BART and GPT3 by a large margin on ROUGE. Furthermore, we introduce Levenshstein Sequential Evaluation metric (LESE) for better evaluation of the structured FAT data and show that it compares exactly with human judgement than existing metrics.
\end{abstract}

\section{Introduction}
Root cause analysis (RCA) in semiconductor industry is the process of discovering the root causes of a failure in order to identify appropriate actions to systematically prevent and solve the underlying issues \citep{abs1999root}. Reliability engineers (experts) in semiconductor industry are usually tasked with carrying out RCA technique known as Failure mode and effects analysis (FMEA). FMEA involves reviewing several components, assemblies, and subsystems to identify potential failure modes in a system and their root causes. This process is done to improve product reliability and quality; cut production cost and reduce defective parts susceptible to failures. Inspection, testing, localization, failure reproduction and documentation are amongst the major steps needed for RCA. Documentation is the most important of all these stages if automation is to be enabled.

Failure Analysis (FA) is a scientific decision making process which aims to discover the root cause for the incorrect behaviour of a product; that is, the failure of a device or a component to meet user expectations \citep{mBazu}.
Failure Analysis (FA) 4.0 addresses a fundamental challenge for the digital industrialized world by ensuring that increasingly complex electronic systems operate with complete reliability and safety in daily use. This is essential in safety-critical applications such as autonomous vehicles \citep{Duan2022, Gultekin2022} and in digitalized industrial production (Industry 4.0) \citep{Behbahani2006200, digitwo, digifa}. The intent of FA 4.0 is therefore to provide innovative AI-based tools and methods to support expert failures analysis during the development and manufacture of electronic components and systems \citep{Ding2021143, Li20221109_fa, Abidi2022, Priyanka2022, Huang2022123}. With its holistic approach spanning chip production \citep{Hu20193201}, assembly and packaging \citep{packaging}, to board and system level, the project’s outcomes will be crucial for the competitiveness of European electronic devices, especially in the demanding automotive and industrial sectors. 

Due to the demanding complexities and time consuming processes involved, failure analysis is carried out manually, driven by single tasks coming from production, reliability testing and field returns. The time-consuming processes does not allow for analysis of combined electrical and material testing and metrology data from along the manufacturing process flow. It is also susceptible to human error, as diagnostic tools are not linked to each other or a central database to provide information for the next steps in a failure analysis workflow. Semi-automating this complex process using AI-based methods would ameliorate some of this human challenges.

Thus, proper documentation allows for a successful FA, helping to identify potential failure modes based on expert experience with similar products and processes. Consequently, a database of failures and their corrective actions are maintained in semiconductor industries today. the \textbf{F}ailure \textbf{R}eporting, \textbf{A}nalysis and \textbf{C}orrective \textbf{A}ction \textbf{S}ystem (\textbf{FRACAS}) is a closed-loop feedback path by which pertinent reliability data is collected, recorded and analyzed during in-house (laboratories) and production/operation to determine where failure concentration in the design of the equipment \citep{fracas, ezukwoke}. The heart of FRACAS is its database management system (DBMS) which classifies failure modes into categories that are essential in identifying the processes in the product (hardware/software) life cycle requiring the most attention for reliability improvement \citet{fracas}. The report obtained from FRACAS DBMS contains information describing the type of failure and origin of detection; a set of analysis (in the form of triplets- Step type, Substep technique and Equipment) proposed to find the failure root cause; conclusion on the outcome of failure analysis. In later sections of this paper, we will refer to failure description as \textbf{pre-triplet} data and the set of analysis as \textbf{triplets} as they are composed of three major keys. Each triplets makes a failure decision and the objective of this paper it to generate a set of $n$-triplets that suites a particular failure description.

The objective of our paper is to present a first of its kind application of a generative sequence-to-sequence language model for structured failure analysis triplet generation given a failure description and a near human judgement evaluation metric to measure the difference between the human and machine generated triplets. The remaining part of Section I introduces the domain of failure analysis decision flow, its formalization and what is meant by pre-trained language model; Section II introduces language models (LM) and pre-trained language model (PLM) for the downstream task of failure analysis triplet generation (FATG); Section III presents the experimental setup; Section IV, the results and finally, the conclusion in Section V.

\subsection{Failure analysis decision flow}
\begin{figure}[!ht]
			\centering
		    \includegraphics[width = 15.3cm]{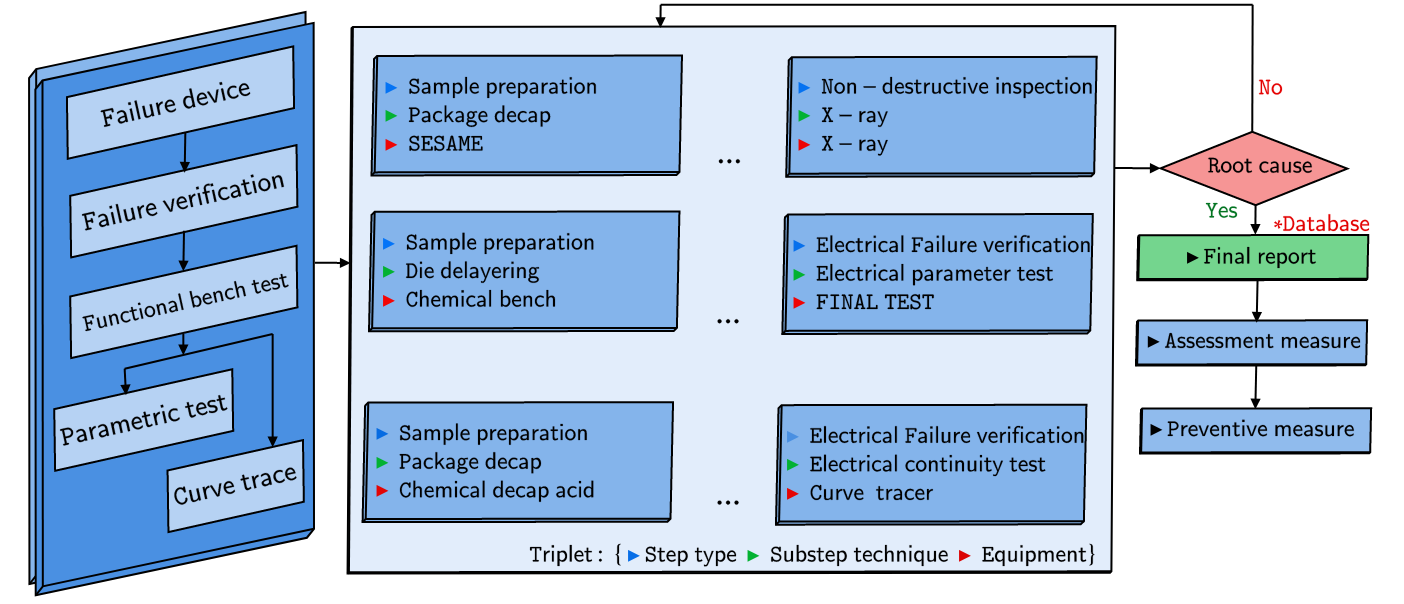}
			\caption{\label{fa_analysis}\textbf{FRACAS}: Structured representation of the failure analysis process. A failure description take indistinct failure analysis flow (path) to finding the root cause of the failure \citep{ezukwoke}. }
\end{figure}

\begin{figure}[!ht]
			\centering
		    \includegraphics[width = 15.3cm]{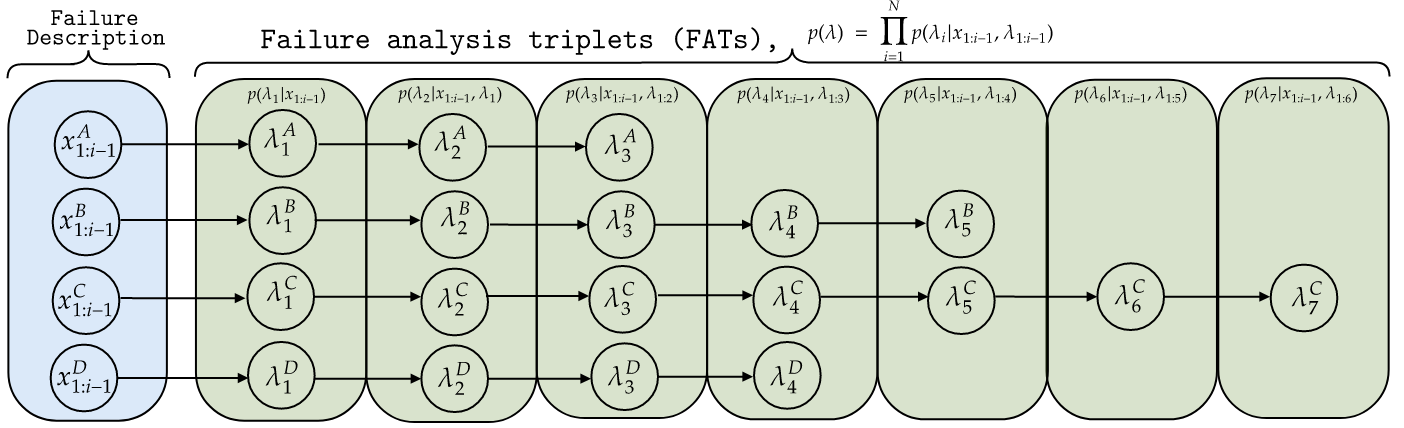}
			\caption{\label{fdr_fa} Failure analysis decision flow graphical formalization. A, B, C and D are different failure description with different failure analysis triplets length. }
\end{figure}

Failure analysis decision flow is a structured database of failure analysis steps performed by industrial experts on a defected device with the expectation of finding the root cause of the failure.
Depending on the cause of the failure and the origin of the failure report, different combinations of equipment are required to find the root cause of a failure. The location of failure analysis lab (FA-lab) is significant in the failure analysis process as the absence of the proper equipment may be lead to repetition of FA, incomplete FA or transfer to different FA-lab. The Figure (\ref{fa_analysis}a) illustrates a typical failure analysis process as seen from the FRACAS perspective. 
An important part of FRACAS is the final report (database) containing details of failure analyses and different combinations of FA. It is important to note that a failure description may have more than one failure analysis path leading to the same finding of the root cause. In fact, one failure description may have two different path with the same sequence of \textbf{step type} and \textbf{substep technique} but only differing in the analysing \textbf{equipment}.\\
\textbf{Failure description (FDR)}. This is a free-text language encoding describing the type and context of failure observed on a microelectronic component or a semiconductor device. Given that free-text from FRACAS report contains lexical and technical jargons, symbolic imputations, abbreviations and numerous language encoding (due to the different geographical locations of the FA-expert); it becomes necessary to preprocess them into tokens understandable by machines using natural language processing techniques before modeling. Similar preprocessing pipeline proposed by \citet{ezukwoke} for FRACAS data (Figure (\ref{fa_analysis}b)) for unsupervised modeling is adopted in this paper.
\\
\textbf{Failure Analysis Triplets (FATs)}. A series of \textbf{Step type}, \textbf{Substep technique} and \textbf{equipment} proposed by a failure analyst in order to find the root cause of a failure and propose a corrective action. The number of failure analysis triplets proposed by a failure analyst (expert) depends on the FDR and the available equipment. Depending on the failure description and available expertise and equipment, different length of FA can be proposed (See Figure \ref{fdr_fa}). The FRACAS database, contains a description, and empty space padded failure analysis associated with the FDR. The padding is based on the longest failure analysis triplets.\\
\textbf{Failure Analysis Triplets Generation (FATG)}. This is a scientific process of generating a series of sequential failure analysis text associated with a failure description. We model the FATG as a sequence-to-sequence data-to-text problem where the input is the FDR (structured tabular data), and the output is a long sequence of a failure analysis triplets. Data-to-text generation is commonly described as a task of automatically producing non-contextual, non-linguistic inputs \citep{gatt_albert}. Early methods apply knowledge-based expert systems \citep{Kukich_karen} for automatically generating natural language reports from computer databases; while recent generation techniques train end-to-end auto-encoding (encoder-decoder) architectures \citep{BahdanauCB14} for similar purpose. Encoder-decoder framework are sequence-to-sequence (Seq2Seq) deep learning-based modeling technique employed in Long-Short-Term Memory (LSTM) \citep{pengfei_liu_rnn, Prajit_Ramachandran, mccann_bryan} and Transformers \citep{attention_is_all} for natural language generation (NLG). In later sections of this paper, we explore language models, in particular, pre-trained language models based on Transformer architecture for FATG. We evaluate the performance of the different models using the popular scoring metrics like BLEU, ROUGE and METEOR. We first formalize our problem in a probabilistic graphical setting before extending it to language modeling.\\
\textbf{Formalization}.
Given failure analysis description (FDR) $\{x_i\}_{i=1}^{N} \in \mathbb{R}^D$, where N is the number of observation and D is the dimension of the preprocessed data, and a set of failure analysis triplets $\{\lambda_i\}_{i=1}^{N} \in \mathbb{R}^M$, where $M$ is the dimension of $\lambda$. We express the FATG as data-to-text problem by defining the input space as a joint space of $x$ and $\lambda$. Let us begin by modeling them individually, the FDR (input space) probability mass function is,
\begin{equation}
    p_{x}(x) = \prod_{i = 1}^N p_{x}(x_i|x_{1:i-1})
\end{equation}
and the failure analysis triplets (FAT),
\begin{equation}
    p_{\lambda}(\lambda) = \prod_{i = 1}^N p_{\lambda}(\lambda_i|\lambda_{1:i-1})
\end{equation}
The joint probability space is given as,
\begin{equation}
    p_{x,\lambda}(x,\lambda) = \prod_{i = 1}^{N} p_{x, \lambda}(\lambda_i|x_{1:i-1}, \lambda_{1:i-1}) \label{x_lambda}
\end{equation}
\textbf{Compact representation}. We present a compact representation for Equation (\ref{x_lambda}) as follows. Let us represent the joint space between failure description $x \in \mathbb{R}^D$ and failure analysis triplets $\lambda \in \mathbb{R}^M$ i.e $\{x_i, \lambda_i\}_{i=1}^N \in \mathbb{R}^K$ as $\Lambda \in \mathbb{R}^K$ , where $K = D+M$ is the dimension of the joint space- a sum of the dimensions of $x$ and $\lambda$ respectively.. We define the compact joint probability space between $x$ and $\lambda$ as,
\begin{equation}
    p(\Lambda) = \prod_{i = 1}^{N} p(\Lambda_i|\Lambda_{1:i-1}) \label{x_lambda_2}
\end{equation}
This compact joint space can be modeled into a likelihood function, $\sum_{i=1}^{N} \log p(\Lambda_i|\Lambda_{1:i-1}; \phi)$ and its parameter, $\phi$ estimated using deep neural network used in training pre-trained language models.
\subsection{Pretrained Language Models (PLMs)}
Interesting developments in deep learning domains resulting from the ability to effectively train models with large parameters has increased rapidly over time. One of the reasons for this is the access to innovative high computing clusters. Large datasets containing millions of observations (with similar or larger number of tokens) is needed to fully train model parameters and prevents overfitting. However, building large-scale labeled datasets can be computational expensive even for a high resource computing cluster for most NLP tasks due to the annotation costs. Pre-trained Model is originally introduced by \citet{NIPS2015_7137debd} for NLP; \citet{pengfei_liu_rnn} as a shared LSTM encoder with Language Model (LM) for multi-task learning (MTL) using fine-tuning; \citet{Prajit_Ramachandran} for unsupervised pre-training of Seq2Seq model and \citet{mccann_bryan} as a deep LSTM encoder from Seq2Seq model with machine learning. 

Since large-scale unlabeled corpora require less computing resources, annotations is not usually needed or  relatively easy to construct. Hence, to leverage the huge unlabeled text data, we can first learn a good representation from unlabelled corpora and then use these representations for other downstream tasks. 
Advantages of such pre-training methods include: \texttt{(i)} Learning universal language representations given their large access to tokens and using them for downstream tasks (such as done here for FATG); \texttt{(ii)} better initialization leading to better generalization on transfer learning; \texttt{(iii)} helps to avoid overfitting on small token data. Recent studies have demonstrated significant performance gains on many NLP tasks with the help of the representation extracted from the PLMs on the large unannotated corpora.

Given that most PLMs have relied greatly on Recurrent Neural Network (RNN) architectures such as Long-Short Term memory (LSTM) \citep{MelisDB18, MelisDB18_2, ConneauKSBB17_emnlp} and Gated Recurrent Neural Network (GRNN) \citep{chaturvedi_etal_2017_story}, it is impossible to ignore their significance for language generation task \citep{Wen_nlp_generation}. RNNs make output predictions at each time step by computing a hidden state vector $H_t$ based on the current input token and the previous states. However, because of their auto-regressive property of requiring previous hidden states vectors $H_{t-n}$ (where $n$ is all hidden states vectors until $t$) to be computed before the current time step, $t$, they require a large amount of time and memory which can be unbearable. The problem with RNN based models is their inability to adapt to long sequence tasks rising from their lack of attention to previously seen sequences; they are computationally demanding and do not benefit from \textit{parallelization}. As a result, Transformer models \citep{attention_is_all} which is built on a Self-attention \citet{attention_is_all} (also an auto-regressive mechanism) is preferred over RNNS. They rely on dot-products between elements of the input sequence to compute a weighted sum \citep{lin_on_attention, karim_ahmed, bahdanau_et_al, kim_et_al_attention} which serves as a critical ingredient in natural language generation tasks. The Table (\ref{tab1}) below shows a list of state-of-the-art pre-trained models built on LSTM and Transformers architectures together with the size of trainable parameters they were trained on.

\begin{table}[!ht]
\centering
\setlength{\tabcolsep}{0.5em} 
{\renewcommand{\arraystretch}{1.7}
\tiny
\begin{tabular}{llll}
\hline
Architecture        &  Model name   & Params      & Project source code                              \\ \hline
LSTM                & LM-LSTM \citep{NIPS2015_7137debd} &      & \texttt{https://github.com/frankcgq105/BERTCHEN}          \\ 
                    & Shared LSTM \citep{pengfei_liu_rnn}&  & \texttt{https://github.com/afredbai/text\_classification} \\ 
                    & ELMo \citep{ELMo} &  & \texttt{https://github.com/dmlc/gluon-nlp}                \\ 
                    & CoVE  \citep{mccann_bryan}&       & \texttt{https://github.com/salesforce/cove}               \\ \hline
Transformer Encoder & BERT (L)  \citep{bert}&  $340$M     & \texttt{https://github.com/google-research/bert}        \\ 
                    & SpanBERT \citep{spanbert}& $340$M  &  \texttt{https://github.com/mandarjoshi90/coref}   \\ 
                    & XLNet (L) \citep{xlnet}&  $360$M     & \texttt{https://github.com/zihangdai/xlnet}    \\ 
                    & RoBERTa \citep{roberta}&  $356$M  & \texttt{https://github.com/facebookresearch/fairseq}  \\ \hline
Transformer Decoder & GPT \citep{gpt}&  $117$M  &  \texttt{https://github.com/huggingface/transformers}    \\ 
                    & GPT-2 \citep{gpt_2}& $1.5$B &  \texttt{https://github.com/huggingface/transformers} \\ 
                    & GPT-3 \citep{gpt_3}& $175$B & \texttt{https://github.com/huggingface/transformers}     \\ \hline
Transformer         & MASS \citep{MASS}& $213$M &   \texttt{https://github.com/microsoft/MASS}    \\ 
                    & BART \citep{BART} &  $380$M     &  \texttt{https://github.com/huggingface/transformers}  \\
                    & T5 \citep{t5} & $11$B & \texttt{https://github.com/huggingface/transformers}  \\ 
                    & XNLG \citep{xnlg} & & \texttt{https://github.com/CZWin32768/XNLG} \\ 
                    & mBART \citep{mBART}& $380$M & \texttt{https://github.com/huggingface/transformers} \\ \hline
\end{tabular}
\caption{\label{tab1}\small Pre-trained Language architectures well established for performing generative task.\\
\textbf{XNLG}: Unspecified parameter size in the original paper. Pre-trained on $10$-hidden encoder layers and $6$-hidden decoder layers with same model parameters as Transformer \citep{attention_is_all}.}
}
\end{table}

In general, pre-trained language models are classified into three taxonomies according to \citet{Qiu_2020} and they include: (i) \textbf{Representation:} based on contextual or non-contextual representation of downstream task; (ii) \textbf{Architectures:} including LSTM, Transformer (standard encoder-decoder transformer), Transformer encoder (only encoder part of Transformer), Transformer decoder (only decoder part of Transformer); (iii) \textbf{Pre-training tasks:} Supervised learning (learning hypothesis/function that maps input to output), Unsupervised learning (finding clusters or latent representation in unlabelled data); Self-supervised learning (combining supervised and unsupervised learning but by first generating training labels automatically).\\
\textbf{Encoder Pre-training}. Encoder pre-training are bi-directional or auto-encoding model that only use encoder during pretraining. Pre-training in this setting is achieved by masking words in the input sentence while training the model to reconstruct. At each stage during pretraining, attention layers can access all the input words. This family of models, for instance BERT \citep{bert} and XLNet \citep{xlnet}, are most useful for tasks that require understanding complete sentences such as sentence classification or extractive question answering. See Table (\ref{tab1}) for other Encoder pre-training examples.\\
\textbf{Decoder Pre-training}. often referred to as auto-regressive models, for example, GPT \citep{gpt_2}, use only the decoder during a pretraining that is usually designed so that the model predict the next words. The attention layers can only access the words positioned before a given word in the sentence, hence, their use for text generation task.

\section{Language Modeling (LM)}
Language Models (LMs), sometimes referred to as Unidirectional or auto-regressive LMs is a probabilistic density estimation approach to solving unsupervised NLP task (such as text generation). Given a text sequence $\{x_i\}_{i=1}^{N} \in \mathbb{R}^D$, its probability density mass $p(x)$ can be expressed as,
\begin{equation}
    p(x) = \prod_{i=1}^{N} p(x_i|x_{0:i-1})
\end{equation}
Where $x_0$ indicates the beginning of the sequence (usually a special token such as $<|\texttt{startoftext}|>$). The conditional probablity can be modeled using a probability distribution over $x_i$ given contextual tokens $x_{0:i-1}$. The contextual tokens can be modeled using neural encoder \texttt{encoder} as follows:
\begin{equation}
    p(x_i|x_{0:i-1}) = \texttt{softmax}(\texttt{encoder}(x_{0:i-1}))
\end{equation}
Note that the \texttt{softmax} function can be any prediction function that fits the expectation and the final hidden layer- \texttt{encoder} can be \texttt{self-masked decoder}. We proceed with training the network with maximum likelihood estimate as follows:
\begin{equation}
    \mathcal{L}_{LM} = \sum_{i=1}^{N} \log p(x_i|x_{0:i-1}; \theta)
\end{equation}
Where $\theta$ is the vector of parameters of the neural network. Other versions of losses are available to training the network depending on the user preference and problem description (See \citet{Qiu_2020}).

\subsection{Transformer Neural Network}
Transformer network \citet{attention_is_all} uses an encoder-decoder architecture with each layer consisting of an attention mechanism called multi-head attention, followed by a feed-forward network. From the source tokens, learned embedding of dimension $d_{model}$ are generated which are then modified by an additive positional encoding (PE).\\
\textbf{Position Embedding} (\textbf{PE}) is computed  using the word position (\textit{pos}) in the sentence and the dimension of the word vector as follows:

\begin{align}
    PE(pos, 2i) = \texttt{sin}(\texttt{pos}/10000^{2i/d_{\texttt{model}}})\\
    PE(pos, 2i+1) = \texttt{cos}(\texttt{pos}/10000^{2i/d_{\texttt{model}}})
\end{align}
The PE corresponds to a sinusoidal wave with wavelength forming a geometric progression between $2\pi - 10000 \cdot 2\pi$ to allow the model learn relative positions.\\
\textbf{Scaled Dot-Product Attention}. A multi-head attention mechanism builds upon scaled dot-product attention, which operates on a query $Q$, key $K$ and a value $V$:
\begin{equation}
    Attention (Q, K, V) = \texttt{softmax}\left(\frac{QK^{T}}{\sqrt{d_k}}\right)V
\end{equation}
Where $d_k$ is the dimension of the key. The input are concatenated in the first layer so that each $(Q, K, V)$ form the word vector matrix. Multi-head attention mechanisms obtain $h$ different representations of $(Q, K, V)$, compute scaled dot-product attention for each representation, concatenate the results, and project the concatenation with a feed-forward layer.
\begin{equation}
    head_i = \texttt{Attention}(QW_i^Q, KW_i^K, VW_i^V)
\end{equation}
\begin{equation}
    MultiHead(Q, K, V) = \texttt{Concat}(head_1, \dots, head_h)W^O
\end{equation}
Where $W_i$ and $W^O$ are the learned projection parameters, $W_i^Q, W_i^K \in \mathbb{R}^{d_{model} \times d_k}$, $W_i^V \in \mathbb{R}^{d_{model} \times d_v}$, $W^O \in \mathbb{R}^{hd_v \times d_{model}}$ and $h$ denotes the number of heads in the multi-attention. $d_v$ is the dimension of the values. \citet{attention_is_all} employs $h=8$ parallel attention layers (heads) and set $d_k = d_v = d_{model}/h = 64$ to achieve the same computational cost as using single-head attention.\\
\textbf{Position-wise Feed-Forward Networks} (\textbf{PFFN}). The component of each layer of the encoder- decoder network is a fully connected feed-forward network applied to each position separately and identically. The authors propose using a two-layered network with a ReLU activation. Given trainable weights $W_1, W_2, b_1, b_2$, the sub-layer is defined as:
\begin{equation}
    FFN(x) = \texttt{max}(0, xW_1 + b_1)W_2 + b_2
\end{equation}
The proposed dimensionality of input and output is $d_{model} = 512$, and the inner-layer has dimensionality $d_{ff} = 2048$ \citep{attention_is_all}.
\begin{figure}
			\centering
		    \includegraphics[width = 15.3cm]{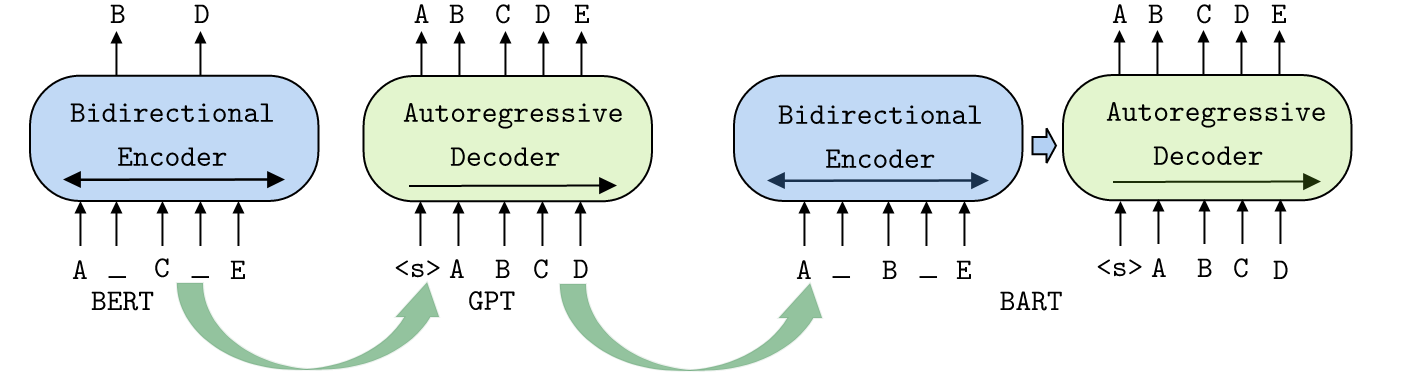}
			\caption{\label{bert_gpt_bart}\textbf{BERT}: Random tokens are replaced with masks, and the document is encoded bidirectionally. Missing tokens are predicted independently. \textbf{GPT}: Tokens are predicted auto-regressively, meaning GPT can be used for generation. However words can only condition on leftward context, so it cannot learn bidirectional interactions. \textbf{BART}: Inputs to the encoder need not be aligned with decoder outputs, allowing arbitrary noise transformations. Here, a document has been corrupted by replacing spans of text with mask symbols. The corrupted document (left) is encoded with a bidirectional model, and then the likelihood of the original document (right) is calculated with an auto-regressive decoder \citep{BART}.}
\end{figure}
\subsection{Generative Pre-trained Transformer (GPT) model} \label{gpt_2_section}
Generative Pre-training model \citep{Radford2018ImprovingLU} is a type of transformer model that uses a multi-layer Transformer decoder \citep{Peter_liu} instead of the encoder-decoder model introduced by \cite{attention_is_all}. 
The model structure begins with training in an unsupervised setting corpus of tokens $\mathcal{U} = \{u_1, \dots, u_n\}$ with a standard likelihood objective to maximize:
\begin{equation}
    L_1(\mathcal{U}) = \sum_i log P(u_i| u_{i-k}, \dots, u_{i-1}; \Theta)
\end{equation}
Where $k$ is the context window and $\Theta$ is the neural network parameters obtained when modeling conditional probability P.
This model applies a multi-headed self-attention operation over the input context tokens followed by position-wise feed-forward layers to produce an output distribution over target tokens as follows:
\begin{equation}
    h_0 = UW_e + W_p
\end{equation}
\begin{equation}
     h_i = \texttt{transformer\_block}(h_{i-1})  \quad \forall i \in [1, n]\end{equation}
\begin{equation}
     P(u) = \texttt{softmax}(h_nW_e^T)
\end{equation}
Where $U = (u_{-k}, \dots, u_{-1})$ is the context vector of tokens, $n$ here is the number of layers while $W_e$ and $W_p$ are the token embedding and position embedding matrix respectively. The results on our data for FATG shows that GPT models outperformed other models and the baseline model by over $100\%$ on BLEU score.

\subsection{BART (Bidirectional Auto-Regressive Transformer)} model\label{bart_2_section}
BART is a Bidirectional Auto-Regressive Transformer trained by (1) corrupting text with an arbitrary noising function, and (2) learning a model to reconstruct the original text \citep{BART}. BART uses a standard Transformer-based neural machine translation architecture (i.e Encoder-Decoder Transformer as seen in Figure \ref{bert_gpt_bart}). The BART loss function, $\mathcal{L}_2(\mathcal{U})$, used to train the pre-trained model is the negative log likelihood of the original training documents:

\begin{equation}
    L_2(\mathcal{U}) = -\sum_i log P(u_i| u_{i-k}, \dots, u_{i-1}; \Theta)
\end{equation}
Note that the corpus of tokens $\mathcal{U} = \{u_1, \dots, u_n\}$ is as seen for GPT while $\Theta$ is the neural network parameters.
\subsection{PLMs for Failure analysis generation domain task}
Despite the overwhelming performance of PLMs on opensource datasets, adapting them to domain specific downstream task can be challenging and sometimes futile. This under-generalization to downstream domain specific task is most often related to transfer learning issues. \textbf{Transfer learning} \citep{transfer_learning} is the ability of a model to transfer the knowledge acquired from a source domain (task) to another domain (often called the target). The inability of PLMs to generalize arise from one or more key factors including: \texttt{(i)} Model architecture not fit for the downstream target domain; \texttt{(ii)} The data distribution of the target domain is far from the source domain on which the PLMs is trained. The later arises, when sufficient domain-specific knowledge for instance, tokens, sentence or characters combination learnt by the PLMs are lacking in the target domain, posing a huge semantic gap.\\
\begin{figure}[!ht]
			\centering
		    \includegraphics[width = 15.3cm]{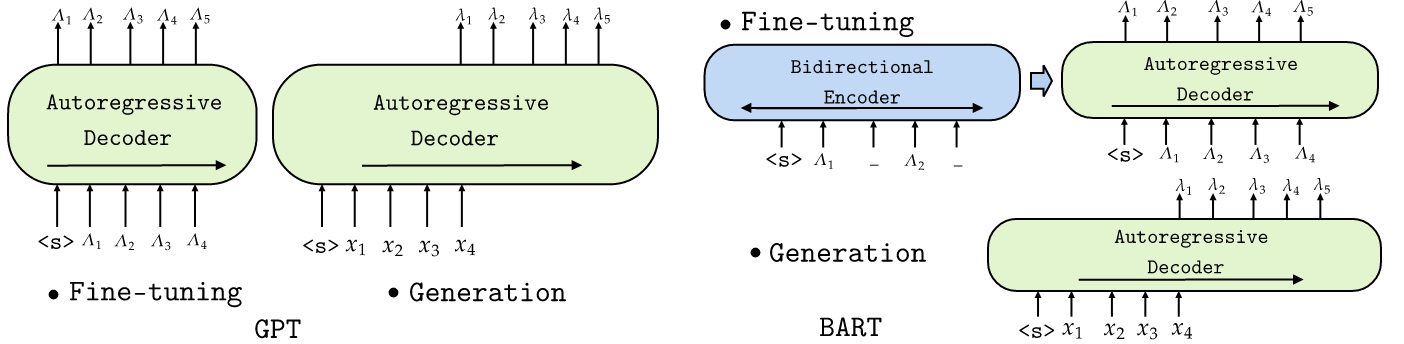}
			\caption{\label{fa_gpt_bart} PLM for Failure analysis triplet generation (FATG). Both GPT and BART architecture are trained on joint input space $\{x, \lambda\} \in \Lambda$. Only the failure description (FDR) is the required input for generation.}
\end{figure}
We apply the weights of the PLMs to our task of failure analysis triplets generation (FATG) using the two transformer models described in previous sections (\ref{gpt_2_section}, \ref{bart_2_section}). Since, there are no labelled instance of FAT datasets, as they are confidential and unavailable from other domains, domain-adaptive intermediate fine-tuning \citep{Phang2018SentenceEO} becomes a challenging task and hence, the poor performance on FATG task reported for GPT3 architecture and BERT (dropped due to lack of meaningful generation). It has been shown that the additional training with general-purpose text corpora (for example, WebText) on the same text generation task can improve the performance on a specific domain \citep{junyi_li_tianyi}. Interestingly, this is the case for GPT2 as it has been trained on WebText corpora, saving us time due intermediate fine-tuning. Results shows outstanding performance for all GPT2 models for the downstream FATG task. \\We fine-tune the PLMs using the vanilla fine-tuning method as seen in DialogGPT \citep{dialog_gpt} for conversational response generation. The input and output data structures used for fine-tuning and generation are illustrated in the Figure (\ref{fa_gpt_bart}) and maintain similar likelihood objective function used for pre-training for the fine-tuning phase.
\section{Experiments}
\textbf{Setup.} Experimentation is carried out on a High performance computing (HPC) cluster  with 80 cores 2 $\times$ Intel Xeon E5-2698 v4 2.20GHz CPU (80 cores), $512$G RAM size and 8 $\times$ Nvidia V100 32G GPU.\\
\textbf{{miniature-GPT}}
For our experiment, we maintain similar properties of the Transformer as proposed by \citet{attention_is_all}. The $d_{model}, d_v, d_k$ are the same. The model is trained for $100$ epochs. The maximum length of sequence is $80$, vocabulary size used is $20$K words, embedding size is $256$. Top-k \citep{holtzman_decoding} decoding is used for generating the FA triplets.\\
\textbf{{Pre-trained GPT-2 \& 3.}} We fine-tune three versions of the GPT-2 (\textbf{gpt2, gpt2-medium ($335$M parameters), gpt2-large ($774$M-$1.5$B parameters)}) for the purpose of failure analysis generation provided by the Huggingface API \footnote{Transformers: \texttt{https://github.com/huggingface/transformers}}. The begin of sequence token, \textbf{bos} is set to $<|\texttt{startoftext}|>$; End of sequence token, \textbf{eos}, $<|\texttt{endoftext}|>$ and padding token, \textbf{pad\_token} is $<|\texttt{pad}|>$. Batch size for training and evaluation is $1$; weight decay is $0.05$ and the number of training epochs is $100$. \textbf{GPT2} is trained on $40$G of WebText data (web pages from outbound links on Reddit) excluding Wikipedia pages; the texts are tokenized using a byte-level version of Byte Pair Encoding (BPE) and contains a vocabulary size of $50257$. The inputs are sequences of $1024$ consecutive tokens. \textbf{OPENAI}-\textbf{GPT3} (\textbf{$175$B parameters and $500$B tokens}) however was trained on BooksCorpus dataset \citep{Zhu2015AligningBA_bookcorpus}, which contains over $7000$ unique unpublished books from a variety of genres including Adventure, Fantasy, and Romance. GPT3 uses similar encoding as GPT2, a bytepair encoding (BPE) vocabulary with $40$K merges.\\
\textbf{{BART Model.}} The \textbf{bos}, \textbf{eos}, \textbf{pad\_token} and tokenization (byte-pair encoding) are same with those used for GPT-2. We use the same pre-training data as \citet{Liu2019RoBERTaAR}, consisting of 160Gb of news, books, stories, and web text.\\
\subsection{Dataset}
We perform experimentation using real failure analysis data from a semiconductor industry, taking into account only successful failure analysis for one year (2019). In order to train the transformer model, we first concatenate all input features along the horizontal axis ($x-axis$) with the triplet data. The size of the data after preprocessing for one year ($2019$) is $5809$. The $10$-input failure description features (see Table \ref{tab2_fdr}) also called \textbf{Expert features} include: \textit{Reference, Subject, Site, Requested activity, Priority level, High confidentiality, Context, Objectives / Work description, Source of failure / request, Source of failure (Detailled)} to train the Transformer model.\\
\begin{table}[!ht]
\centering
\setlength{\tabcolsep}{0.5em} 
{\renewcommand{\arraystretch}{1.7}
\scriptsize
\begin{tabular}{lll}
\hline
\textbf{Feature}                      & \textbf{Description}                                                                                                              & \textbf{Example}                         \\ \hline
Reference                    & \begin{tabular}[l]{@{}l@{}}Concatenated tokens describing the failure \\ analysis team, FA-lab, year and ID\end{tabular} & PTM BOUSKOURA \_18\_04655       \\ \hline
Subject                      & \begin{tabular}[l]{@{}l@{}}A unique subject particular to the fault expert\\  desire to analyse\end{tabular}             & Load failure                    \\ \hline
Site                         & Failure analysis lab site                                                                                                & Grenoble                        \\ \hline
Requested activity           & \begin{tabular}[l]{@{}l@{}}Major fault analysis requested by expert or\\  client\end{tabular}                            & Failure Analysis                \\ \hline
Priority level               & Level of priority given to failure device                                                                                & P1, P2 or P3                    \\ \hline
High confidentiality         & Confidentiality level                                                                                                    & Yes or No                       \\ \hline
Context                      & Context reason for the failure analysis                                                                                  & Non-operational Heat Soak Fail  \\ \hline
Objective/Work description   & Objective of the failure analysis                                                                                        & Check possible cause of failure \\ \hline
Source of failure/Request    & Entity that identified the failure                                                                                       & Customer Complaint              \\ \hline
Source of failure (Detailed) & Details on the failure source                                                                                            & Reliability Failure             \\ \hline
\end{tabular}
\caption{\label{tab2_fdr}\small Brief description of the input failure description (FDR) variables and their corresponding examples.}
}
\end{table}
The features are concatenated along the horizontal axis with the FATs for Modeling. FATs has a dimension of $\mathbb{R}^{69}$ with the longest FA having $23$ triplets.
We preprocess FDR, $x$ according to the scheme presented in \citet{ezukwoke} and vectorize the joint space $\{x, \lambda\}$ using a byte-level version of Byte Pair Encoding (BPE) from GPT2.

\subsection{Evaluation metric}
\textbf{BLEU} (Bilingual Evaluation Understudy) is a context-free precision-based metric for evaluating the quality of text which has been machine-translated from one natural language to another \citep{bleuscore, bleuscore_2} and has also been adopted for dialog generation task \citep{nlg_metric}. It is a precision-based metric that computes the n-gram overlap between the reference (original) and its hypothesis. In particular, BLUE is the ratio of the number of overlapping $n$-grams to the total number of $n$-grams in the hypothesis. It is formally defined as,
\begin{equation}
    BLEU-N = \text{BP} \cdot exp\left(\sum_{n=1}^N \mathbf{w_n} \log Precision_n\right)
\end{equation}
$\mathbf{w_n}$ is the weights of the different $n-grams$ and BP known as the \textbf{brevity penalty} is used to discourage short meaningless hypothesis,
\begin{equation}
    BP = \begin{cases}
                1, & \text{if $|\texttt{hyp}|$} >|\texttt{ref}|\\
                e^{\left(1- \frac{|\texttt{ref}|}{|\texttt{hyp}|}\right)} & \text{otherwise}
             \end{cases}
\end{equation}
Where $|\texttt{ref}|$ and $|\texttt{hyp}|$ are the length of the reference and hypothesis respectively. Precision is defines as,
\begin{equation}
    Precision_n = \frac{\sum_{p_n \in \texttt{hyp}} \sum_{\texttt{n-gram} \in p_n} Count_{clip}(\texttt{n-gram})}{\sum_{p_n \in \texttt{hyp}} \sum_{\texttt{n-gram} \in p_n} Count(\texttt{n-gram})}
\end{equation}
Where $p_n$ is a subset of $n$-tokens in the hypothesis tokens and $\text{Count}_{clip}(n-gram)$ is the maximum number of times the given $n$-gram appears in any one of the corresponding reference sentences.
In our case, we use it to compare the original expert triplet failure analysis to the Transformer generated triplets without respect to context. The average BLEU score (Avg-BLEU) is computed on a one-to-one basis (Expert triplets-to-Transformer triplets).\\
\textbf{ROUGE} (Recall-Oriented Understudy for Gisting Evaluation) \citep{rouge} is a recall-based metric similar to BLEU-N in counting the n-gram matches between the hypothesis and reference. It is computed as follows:
\begin{equation}
    ROUGE-N = \frac{\sum_{s_r \in \texttt{ref}} \sum_{\texttt{n-gram} \in s_r} Count_{match}(\texttt{n-gram})}{\sum_{s_r \in \texttt{ref}} \sum_{\texttt{n-gram} \in s_r} Count(\texttt{n-gram})}
\end{equation}
Where \texttt{ref} is the reference text and $s_r$ is the hypothesis summary. ROUGE-L is the F-measure of the longest common subsequence (LCS) between a pair of sentences. LCS(p, r) is the common subsequence in p and r with maximum length. ROUGE-L is computed as follows:
\begin{equation}
    ROUGE-L = \frac{(1+\beta^2) R_{LCS}P_{LCS}}{R_{LCS} + \beta^2 P_{LCS}}
\end{equation}
Where Precision,
\begin{equation}
    P_{LCS} = \frac{|LCS(\texttt{ref}, \texttt{hyp})|}{\text{Num. words in hypothesis}}
\end{equation}
and Recall,
\begin{equation}
    R_{LCS} = \frac{|LCS(\texttt{ref}, \texttt{hyp})|}{\text{Num. words in reference}}
\end{equation}
ROUGE-L is a matching and credible metric for the downstream task of failure analysis triplets generation task, as it does not take into account, contextual meaning but rather, word order and their maximum length. Its F-measure is highlighted to show its significance. The weight parameter $\beta^2$ emphasizes the precision as much as the recall.\\
\textbf{METEOR} proposed by \citet{meteor} addresses the major drawback of BLEU including, inability to account for recall and inflexible $n$-gram matching by proposing an F-measure with flexible n-gram matching criteria. The F-measure is computed as follow:
\begin{equation}
    F-score = \frac{10 \text{Precision}\times \text{Recall}}{\text{Recall} + 9\text{Precision}}
\end{equation}
Where
\begin{equation}
    Precision = \frac{\text{Num. mapped unigrams}}{\text{Num. unigrams in candidate}}
\end{equation}
and
\begin{equation}
    Recall = \frac{\text{Num. mapped unigrams}}{\text{Num. unigram in reference}}
\end{equation}
METEOR tends to reward longer contiguous matches using a penalty term known as \textit{fragmentation penalty} \citep{meteor}.
\subsubsection{\textbf{LESE}: LEvenshtein Sequential Evaluation metric}
For better understanding of the sequential generation of FATs and to address the drawback of ROUGE, we propose \textbf{LE}venshtein  \textbf{S}equential \textbf{E}valuation (\textbf{LESE}) metric. LESE is an $n$-gram Levenshtein distance \citep{Levenshtein} based metric used for measuring the similarity between two sequences by computing the $n$-gram edits (insertions, deletions or substitutions) required to change one sequence into another. It is unbiased in the computation of precision and recall as it takes into account the total number of $n$-grams as the denominator, in the computation of the precision and recall rather than unigrams. Given a reference sequence, \texttt{ref}$ = a_1, \dots, a_m$ and a hypothesis sequence, \texttt{hyp}$ = b_1, \dots, b_n$, where $m$ and $n$ are the length of \texttt{ref} and \texttt{hyp} respectively, the nonsymmetric $n$-gram Levenshtein distance (\textbf{Lev}) is given by the recurrence,
\begin{align}
    Lev_{(i,j)}(\texttt{r,h}) = 
    \begin{cases}
    Lev_{(i,0)} = \sum_{k=1}^{i} w(a_k) \qquad \text{for }1\leq i\leq m \\
    Lev_{(0,j)} = \sum_{k=1}^{j} w(b_k) \qquad \text{for }1\leq j\leq n \\
    Lev_{(i-1,j-1)} \qquad \qquad \text{for }a_{i:i+n-\text{gram}}= b_{j:j+n-\text{gram}} \\
    1+ min \begin{cases}
    Lev_{\color{red}d}(i-1,j) \\
    Lev_{\color{red}i}(i,j-1) \text{ for }a_{i:i+n-\text{gram}} \not = b_{j:j+n-\text{gram}} \\
    Lev_{\color{red}s}(i-1,j-1)\\
    \end{cases}
    \end{cases}
\end{align}
$Lev_{(m,n)}(\texttt{r,h}) = Lev_{(m,n)}(\texttt{r,h})//\texttt{n-gram}$ is the $n$-gram Levenshtein distance; \texttt{r} and \texttt{h} refer to the reference and hypothesis sequence while $Lev_{\color{red}d}$, $Lev_{\color{red}i}$ and $Lev_{\color{red}s}$ denote the deletion, insertion and substitution operations. The smaller the  Levenshstein distance between two sequences, the higher their similarity, hence, \texttt{Lev(ref, hyp) = 0} when \texttt{hyp = ref}. The LESE-N (F1-score) is computed thus,
\begin{equation}
    \textbf{LESE-N} = \frac{(1+\beta^2)P_{\texttt{Lev}}R_{\texttt{Lev}}}{\beta^2P_{\texttt{Lev}}+R_{\texttt{Lev}}} \label{lese_eq}
\end{equation}
 The value of $\beta$ in Equation (\ref{lese_eq}) is $1$ for all experiments with Precision and Recall defined as follow,
\begin{align}
    \begin{cases}
    P_{\texttt{Lev}} = \texttt{max}\Bigl\{0, \left|\frac{|\texttt{max}(\texttt{|r|}_{\texttt{n-gram}}, \texttt{|h|}_{\texttt{n-gram}}) - \texttt{Lev}_{(m,n)}(\texttt{r, h})|}{\texttt{|n-gram in hypothesis|}}\right|\Bigl\} \\
    R_{\texttt{Lev}} = \texttt{max}\Bigl\{0, \left|\frac{|\texttt{max}(\texttt{|r|}_{\texttt{n-gram}}, \texttt{|h|}_{\texttt{n-gram}}) - \texttt{Lev}_{(m,n)}(\texttt{r, h})|}{\texttt{|n-gram in reference|}}\right|\Bigl\} \\
    \end{cases}  \label{lev_pr}
\end{align}
The numerators in Equation (\ref{lev_pr}) is the length between the  maximum $n$-grams in \texttt{|ref|} and \texttt{|hyp|}, and the Levenshstein distance, \texttt{Lev(ref, hyp)}. Maximum is taken to avoid numerical problem that arise from empty sequence for either reference or hypothesis (as observed for FATG) or sometimes both.

When compared to human evaluation, LESE-N performs comparatively similar, especially LESE-3 for FATG compared to other metric used for quantitative evaluation.
\subsection{Decoding method}
\textbf{Top-$k$ Sampling}. Given a conditional probability distribution $p(x_t|x_{0:t-1})$, it top-$k$ vocabulary \{$V: V^k \subset V$\} is defined as the set of size -$k$ which maximizes the conditional probability $\sum_{x \in V^{k}} p(x_t|x_{0:t-1})$. At each time step, the top-$k$ possible next tokens are sampled according to their relative probabilities. The conditional distribution can be re-scaled thus,
\begin{equation}
    p^{\prime}(x_i|x_{0:i-1}) = \begin{cases}
    \frac{p(x_i|x_{0:i-1})}{\sum_{x \in V^{k}} p(x_i|x_{0:i-1})}, & \text{if $x_i$}  \in V^{k}\\
    0, & \text{otherwise}
    \end{cases}
\end{equation}
and then sampling on the re-scaled probabilities.
\\
\textbf{Nucleus Sampling}\footnote{Nucleus sampling: \texttt{https://github.com/ari-holtzman/degen}} \citep{nucleus_sampling}. Given a conditional probability distribution $p(x_t|x_{0:t-1})$, the top-$p$ vocabulary \{$V: V^p \subset V$\} is denoted by,
\begin{equation}
    \sum_{x \in V^{p}} p(x_i|x_{0:i-1})\ge p_i
\end{equation}
Where $p_t$ is a user defined threshold taking values between $[0, 1]$. otherwise known as top-$p$, nucleus sampling is a stochastic decoding sampling method that uses the shape of a probability distribution to determine the set of tokens to sample from. It avoids text degeneration by truncating the tail of the probability distribution. Furthermore, it resolves the challenge of flat and peaked-distribution faced when using top-$k$ by re-scaling the distribution $p(x_t|x_{0:t-1})$ from which the next token is sampled as,
\begin{equation}
    p^{\prime}(x_i|x_{0:i-1}) = \begin{cases}
    \frac{p(x_i|x_{0:i-1})}{\sum_{x \in V^{p}} p(x_i|x_{0:i-1})}, & \text{if $x_i$}  \in V^{p}\\
    0, & \text{otherwise}
    \end{cases}
\end{equation}
The size of the sampling set will adjust dynamically based on the shape of the probability distribution at each time-step \citep{nucleus_sampling}.
\section{Results}
\subsection{Quantitative Evaluation} \label{quant_eval}
We evaluate the performance of PLMs for Failure analysis generation (FATG). Evaluating the generative power of a model requires an equally optimized decoding method. We combine top-$p = 0.95$ and top-$k = 10$ with a normalizing temperature value of $1.9$ for decoder sampling of pretrained \textbf{BART}, \textbf{GPT2}, \textbf{GPT2-Medium}, \textbf{GPT2-Large} and \textbf{OPENAI-GPT3} models. Top-$k$ (with $k=10$) performs well for the generative decoder sampling of \textbf{mini-GPT}.
\\
\begin{table}[!ht]
\centering
\setlength{\tabcolsep}{0.5em} 
{\renewcommand{\arraystretch}{1.2}
\tiny
\renewcommand{\tabcolsep}{1.7pt}
\begin{tabular}{c
>{\columncolor[HTML]{EFEFEF}}c c
>{\columncolor[HTML]{EFEFEF}}c cccccc
>{\columncolor[HTML]{DAE8FC}}c 
>{\columncolor[HTML]{DAE8FC}}c 
>{\columncolor[HTML]{DAE8FC}}c 
>{\columncolor[HTML]{EFEFEF}}c 
>{\columncolor[HTML]{DAE8FC}}c 
>{\columncolor[HTML]{DAE8FC}}c 
>{\columncolor[HTML]{DAE8FC}}c 
>{\columncolor[HTML]{EFEFEF}}c }
\hline
                        & \cellcolor[HTML]{EFEFEF}                         &                                       & \cellcolor[HTML]{EFEFEF}                       & \multicolumn{3}{c}{ROUGE-1}                                                                                                                   & \multicolumn{3}{c}{ROUGE-L}                                                                                                                   & \multicolumn{3}{c}{\cellcolor[HTML]{DAE8FC}LESE-1}                                                                    & \cellcolor[HTML]{EFEFEF}                        & \multicolumn{3}{c}{\cellcolor[HTML]{DAE8FC}LESE-3}                                                                    & \cellcolor[HTML]{EFEFEF}                        \\ \cline{5-13} \cline{15-17}
\multirow{-2}{*}{Model} & \multirow{-2}{*}{\cellcolor[HTML]{EFEFEF}BLUE-1} & \multirow{-2}{*}{BLEU-3}              & \multirow{-2}{*}{\cellcolor[HTML]{EFEFEF}MET.} & Prec.                                 & Rec.                                  & \cellcolor[HTML]{EFEFEF}F1                                    & Prec.                                 & Rec.                                  & \cellcolor[HTML]{EFEFEF}F1                                    & Prec.                                 & Rec.                                  & F1                                    & \multirow{-2}{*}{\cellcolor[HTML]{EFEFEF}Lev-1} & Prec.                                 & Rec.                                  & F1                                    & \multirow{-2}{*}{\cellcolor[HTML]{EFEFEF}Lev-3} \\ \hline
mini-GPT                & 11.54                                            & 7.51                                  & {\color[HTML]{3531FF} \textbf{34.61}}          & 11.22                                 & 16.12                                 & \cellcolor[HTML]{EFEFEF}12.63                                 & 10.19                                 & 14.79                                 & \cellcolor[HTML]{EFEFEF}11.52                                 & 8.11                                  & 8.57                                  & 7.11                                  & 46.69                                           & 0.38                                  & 0.27                                  & 0.30                                  & 16.0                                           \\ \hline
BART$^\dagger$                    & 6.14                                             & 4.71                                  & 9.23                                           & 11.24                                 & 10.79                                 & \cellcolor[HTML]{EFEFEF}10.16                                 & 10.40                                 & 10.06                                 & \cellcolor[HTML]{EFEFEF}9.43                                  & 5.26                                  & 4.04                                  & 4.08                                  & 42.71                                           & 1.68                                  & 1.29                                  & 1.28                                  & {\color[HTML]{3531FF} \textbf{14.0}}           \\ \hline
GPT3$^\dagger$                    & 6.10                                             & 2.30                                  & 30.69                                          & 7.07                                  & 13.49                                 & \cellcolor[HTML]{EFEFEF}8.84                                  & 6.29                                  & 12.22                                 & \cellcolor[HTML]{EFEFEF}7.91                                  & 4.10                                  & 9.83                                  & 5.46                                  & 82.16                                           & 0.41                                  & 0.63                                  & 0.42                                  & 28.0                                           \\ \hline
GPT2$^\dagger$                    & 22.18                                            & 17.39                                 & 29.71                                          & 30.25                                 & 33.79                                 & \cellcolor[HTML]{EFEFEF}29.67                                 & 28.35                                 & 31.62                                 & \cellcolor[HTML]{EFEFEF}27.75                                 & 22.04                                 & 23.99                                 & 20.97                                 & {\color[HTML]{3531FF} \textbf{42.26}}           & 11.03                                 & 11.99                                 & 10.49                                 & 15.0                                           \\
GPT2-M$^\dagger$                  & 22.15                                            & 17.32                                 & 30.38                                          & 29.89                                 & 34.36                                 & \cellcolor[HTML]{EFEFEF}29.69                                 & 27.97                                 & 32.20                                 & \cellcolor[HTML]{EFEFEF}27.78                                 & {\color[HTML]{3531FF} \textbf{21.97}} & 24.81                                 & 21.21                                 & 43.28                                           & {\color[HTML]{3531FF} \textbf{11.10}} & {\color[HTML]{3531FF} \textbf{12.56}} & {\color[HTML]{3531FF} \textbf{10.74}} & 15.0                                           \\
GPT2-L$^\dagger$                  & {\color[HTML]{3531FF} \textbf{22.46}}            & {\color[HTML]{3531FF} \textbf{17.60}} & {\color[HTML]{000000} 30.79}                   & {\color[HTML]{3531FF} \textbf{29.73}} & {\color[HTML]{3531FF} \textbf{34.77}} & \cellcolor[HTML]{EFEFEF}{\color[HTML]{3531FF} \textbf{29.82}} & {\color[HTML]{3531FF} \textbf{27.86}} & {\color[HTML]{3531FF} \textbf{32.63}} & \cellcolor[HTML]{EFEFEF}{\color[HTML]{3531FF} \textbf{27.93}} & 21.88                                 & {\color[HTML]{3531FF} \textbf{24.91}} & {\color[HTML]{3531FF} \textbf{21.25}} & 43.41                                           & 11.04                                 & {\color[HTML]{3531FF} \textbf{12.56}} & 10.73                                 & 16.0                                           \\ \hline
\end{tabular}
\caption{\label{eval_result}\small Model comparison on BLEU \citep{bleuscore}, ROUGE-1 \& L \citep{rouge} and METEOR (MET.) \citep{meteor} scores (\%). Higher values (in {\color{blue}{\textbf{bold-blue}}}) is preferred for all metric except \textbf{Lev} (average $n$-gram Levenshstein distance). GPT2-L (Large) and GPT2-M (Medium) are both competitive in performance with Large, averagely performing better on BLEU, ROUGE and LESE scores compared to baseline (mini-GPT). GPT2-M performs best on LESE-3 triplet score, closely followed by GPT2-L. Prec. Rec; and F1 denote Precision, Recall and F1-score respectively. \{Model\}$^{\dagger}$: Pre-trained model.}
}
\end{table}
\comment{
\begin{figure}[t]
			\centering
			\subfloat[\centering mini-GPT]{{\includegraphics[width = 2.58cm]{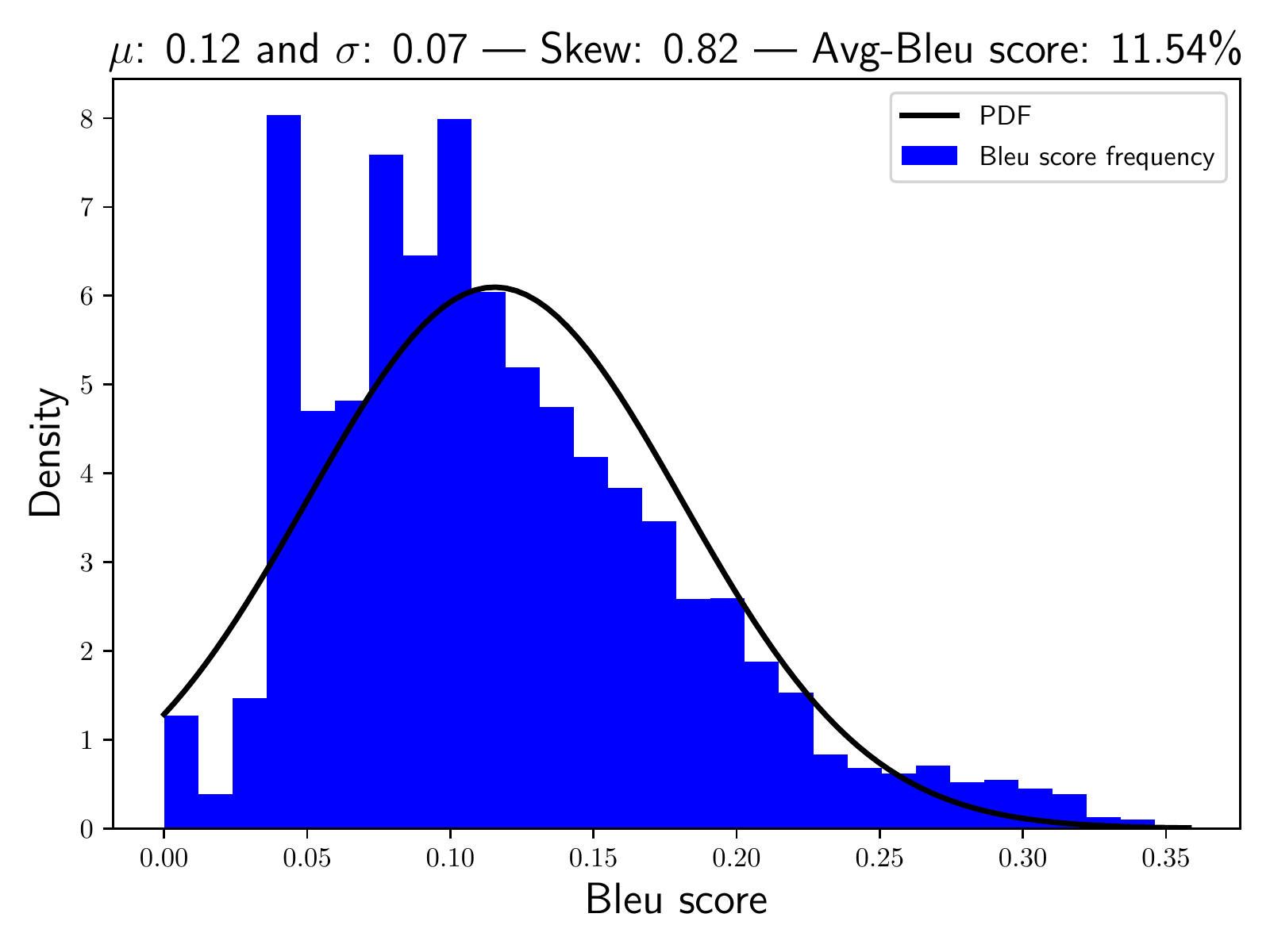}}}
			\subfloat[\centering PT-BART]{{\includegraphics[width = 2.30cm]{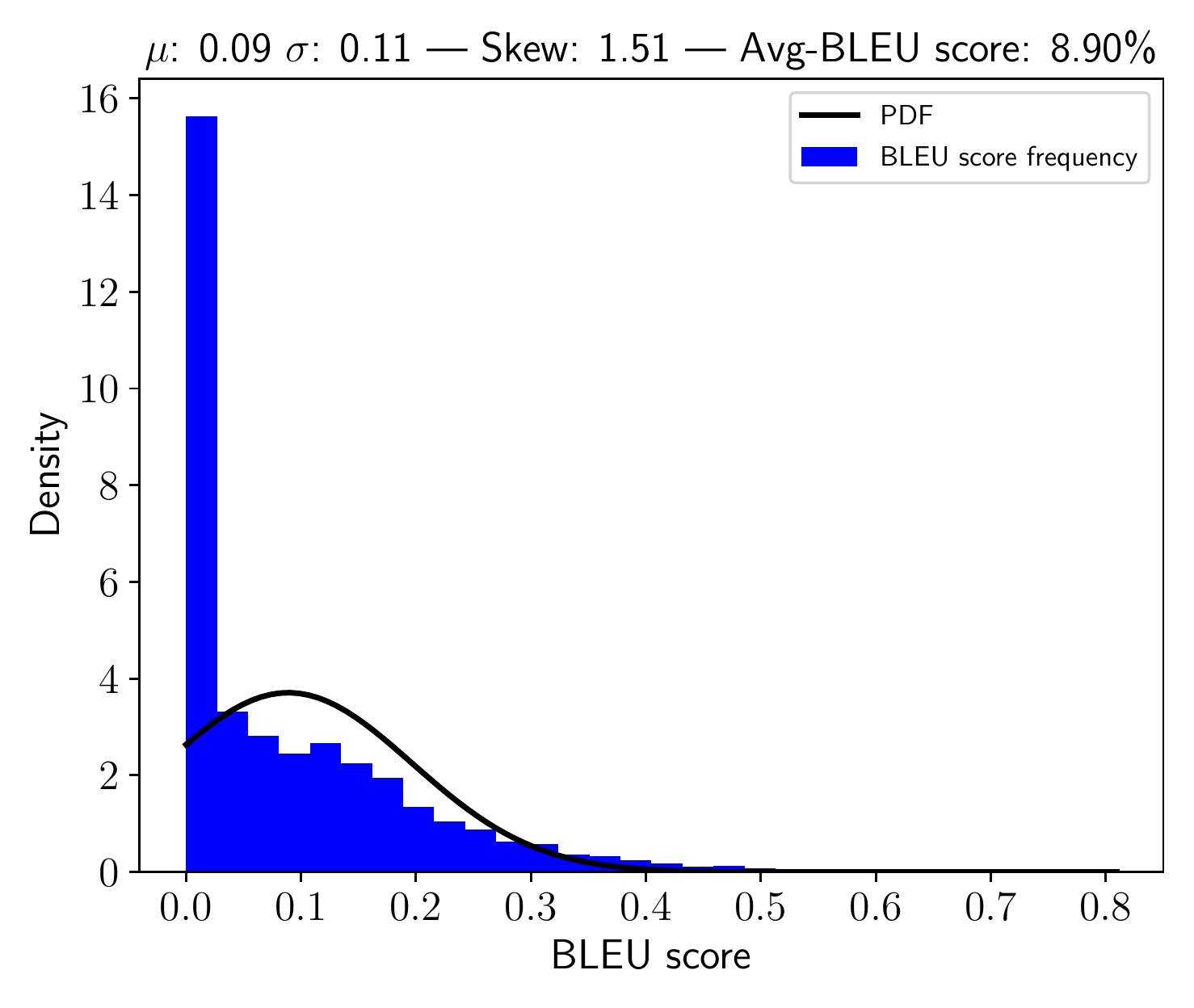}}}
			\subfloat[\centering PT-GPT3]{{\includegraphics[width = 2.40cm]{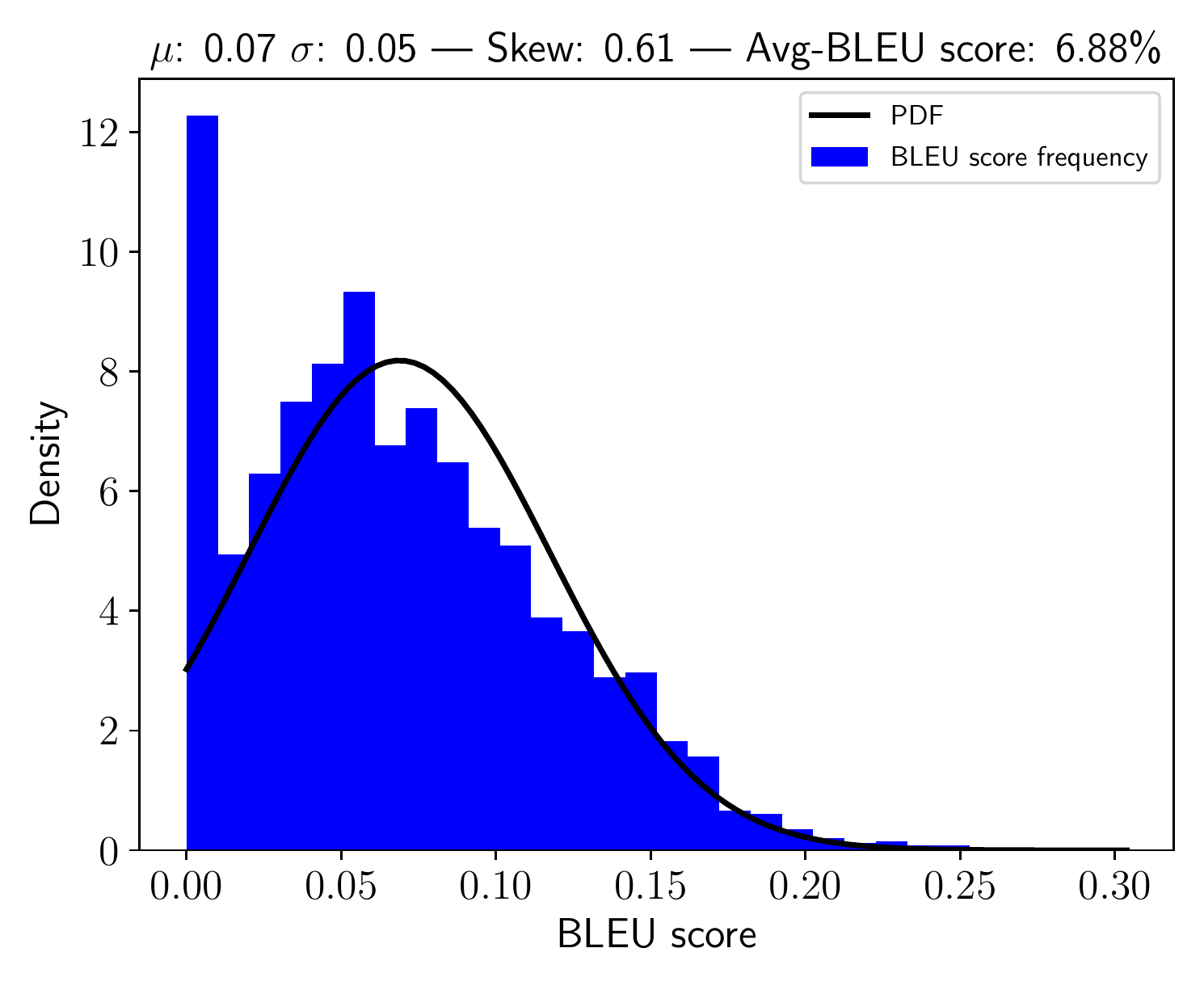}}}
		    \subfloat[\centering PT-GPT2]{{\includegraphics[width = 2.38cm]{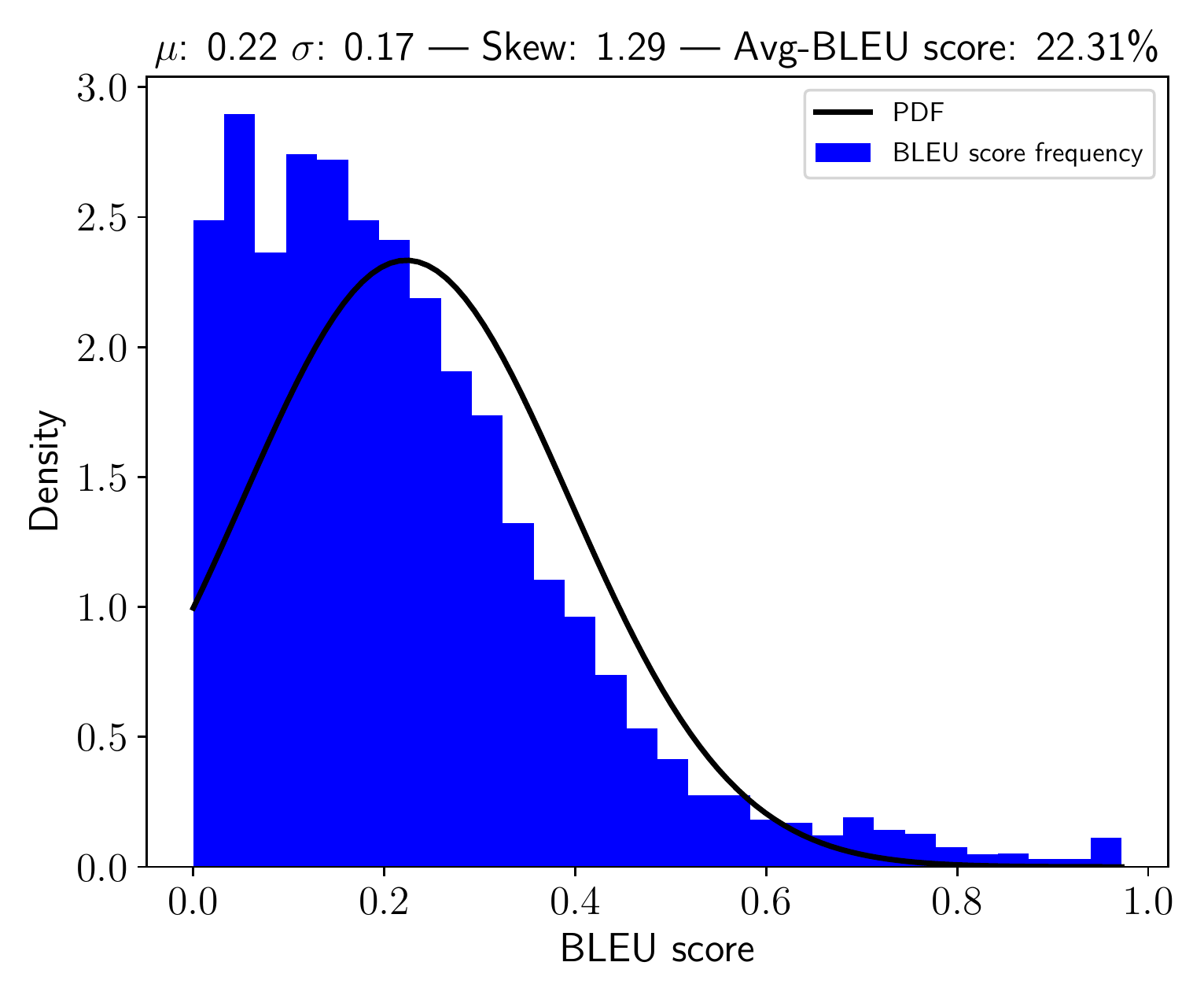}}}
			\subfloat[\centering PT-GPT2 Medium]{{\includegraphics[width = 2.38cm]{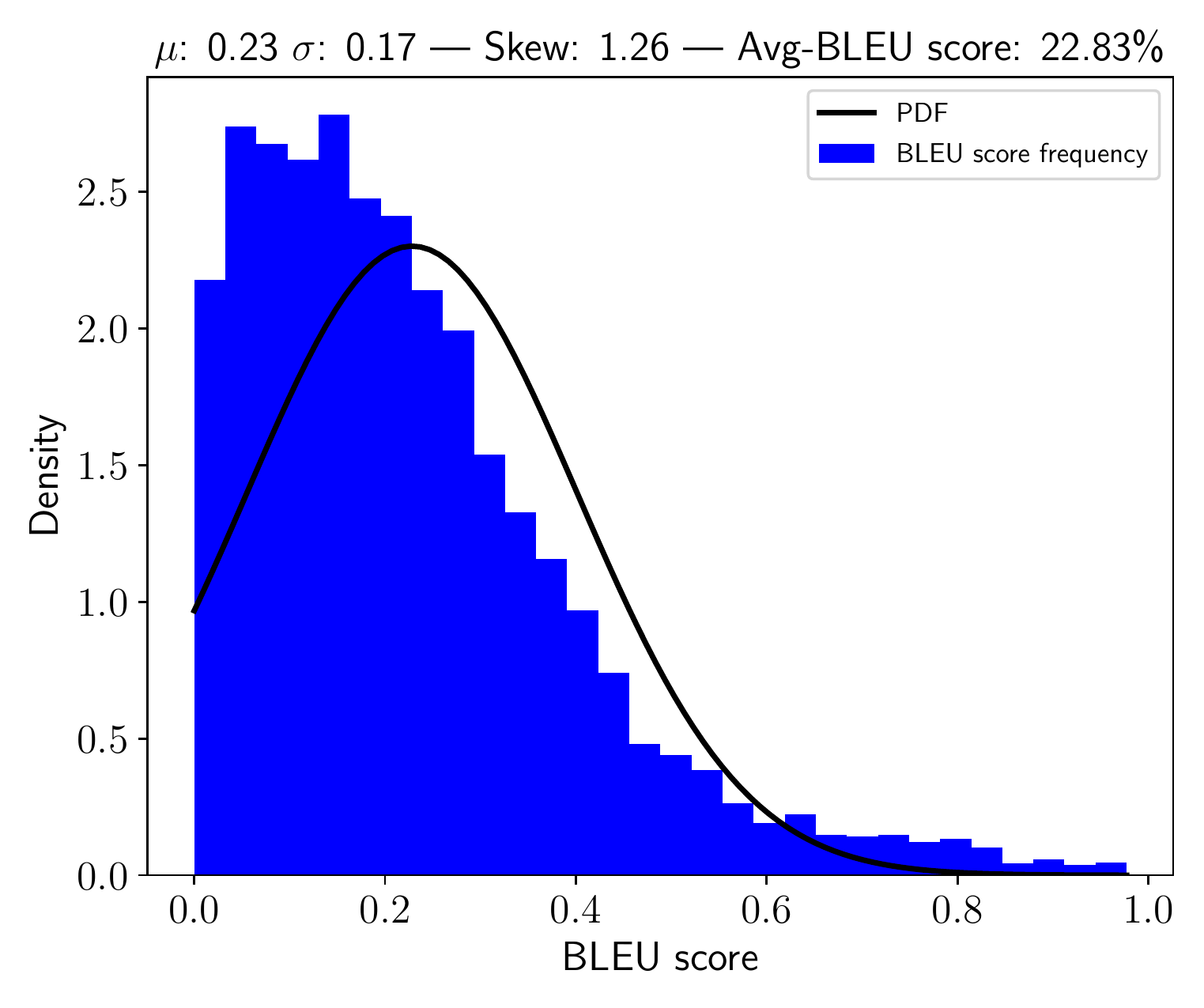}}}
			\subfloat[\centering PT-GPT2 Large]{{\includegraphics[width = 2.48cm]{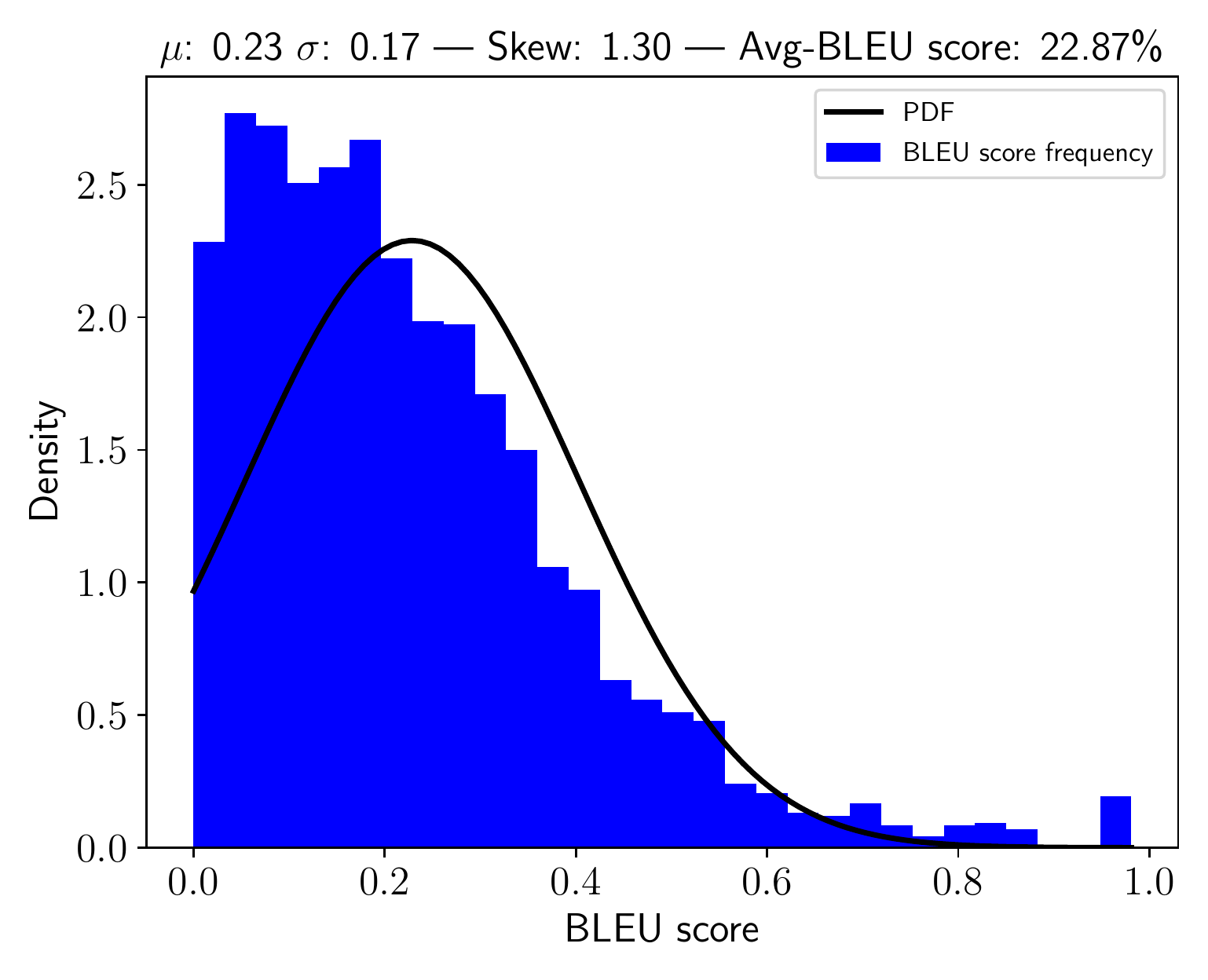}}}
			\caption{\label{blue_score}\small Density distribution of the BLEU scores for the different models. GPT2 base, Medium and Large are positively skewed and have similar density distribution. }
\end{figure}
}
Results (as seen on Table \ref{eval_result}) indicates GPT2-Model trained on the Webtext dataset performs considerably better than the baseline (mini-GPT), GPT3 and BART models. The best peforming GPT2 (GPT2 Large) outperforms the base model by $49\%$ on the BLEU metric; $73\%$ better than BART and GPT3 for FATG. The LESE metric evidently shows that GPT2, GPT2 Medium and GPT2 Large is trained on correlating domain knowledge related to failure analysis and reliability engineering, hence, the adaptive transfer of knowledge for failure analysis triplets generation.
\comment{
\begin{table}[!ht]
    \scriptsize
    \centering
    \setlength{\tabcolsep}{0.2em} 
    {\renewcommand{\arraystretch}{1.2}
    \centering
    \begin{tabular}{ccccc}
        \hline
        \multicolumn{5}{c}{{\begin{tabular}[c]{@{}l@{}} \textbf{F1-score} \end{tabular}}} \\ \hline
        Model         & \cellcolor[HTML]{EFEFEF} \textbf{ROUGE-1} & \textbf{ROUGE-2} & \cellcolor[HTML]{EFEFEF}\textbf{ROUGE-L}   & \textbf{METEOR} \\ \hline
        \textbf{PT-GPT2}           &  \cellcolor[HTML]{EFEFEF} &    &    \cellcolor[HTML]{EFEFEF}  &          \\ 
        \textbf{PT-GPT2 Medium} &    \cellcolor[HTML]{EFEFEF}   &  \color{blue}\ding{51} &  \cellcolor[HTML]{EFEFEF}  &     \color{blue}\ding{51}   \\ 
        \textbf{PT-GPT2 Large}    & \cellcolor[HTML]{EFEFEF}\color{blue}\ding{51} &     &    \cellcolor[HTML]{EFEFEF}\color{blue}\ding{51}  &      \\ \hline
    \end{tabular}
    \caption{\small F1-score comparison for best performing model for ROUGE metric and METOER}
    \label{f1_score_marked}
    }
\end{table}
}
\begin{table}[!ht]
\centering
{\renewcommand{\arraystretch}{1.2}
\tiny
\renewcommand{\tabcolsep}{1.7pt}
\begin{tabular}{cc
>{\columncolor[HTML]{EFEFEF}}c c
>{\columncolor[HTML]{EFEFEF}}c cccccc
>{\columncolor[HTML]{DAE8FC}}c 
>{\columncolor[HTML]{DAE8FC}}c 
>{\columncolor[HTML]{DAE8FC}}c 
>{\columncolor[HTML]{EFEFEF}}c 
>{\columncolor[HTML]{DAE8FC}}c 
>{\columncolor[HTML]{DAE8FC}}c 
>{\columncolor[HTML]{DAE8FC}}c 
>{\columncolor[HTML]{EFEFEF}}c }
\hline
                         &                        & \cellcolor[HTML]{EFEFEF}                                            &                                             & \cellcolor[HTML]{EFEFEF}                                            & \multicolumn{3}{c}{ROUGE-1}                                                         & \multicolumn{3}{c}{ROUGE-L}                                                         & \multicolumn{3}{c}{\cellcolor[HTML]{DAE8FC}LESE-1}                                                                                  & \cellcolor[HTML]{EFEFEF}                        & \multicolumn{3}{c}{\cellcolor[HTML]{DAE8FC}LESE-3}                                                          & \cellcolor[HTML]{EFEFEF}                        \\ \cline{6-14} \cline{16-18}
\multirow{-2}{*}{Weight} & \multirow{-2}{*}{Year} & \multirow{-2}{*}{\cellcolor[HTML]{EFEFEF}BLUE-1}                    & \multirow{-2}{*}{BLEU-3}                    & \multirow{-2}{*}{\cellcolor[HTML]{EFEFEF}MET.}                      & Prec. & Rec.  & \cellcolor[HTML]{EFEFEF}F1                                          & Prec. & Rec.  & \cellcolor[HTML]{EFEFEF}F1                                          & Prec.                         & Rec.                          & F1                                                                  & \multirow{-2}{*}{\cellcolor[HTML]{EFEFEF}Lev-1} & Prec.                         & Rec.                          & F1                                          & \multirow{-2}{*}{\cellcolor[HTML]{EFEFEF}Lev-3} \\ \hline
$w^{\dagger}$                        & 2019                   & 22.15                                                               & 17.32                                       & 30.38                                                               & 29.89 & 34.36 & \cellcolor[HTML]{EFEFEF}29.69                                       & 27.97 & 32.20 & \cellcolor[HTML]{EFEFEF}27.78                                       & 21.97                         & 24.81                         & 21.21                                                               & 43.28                                           & 11.10                         & 12.56                         & 10.74                                       & 15.0                                           \\ \hline
{$w_{2019}$}                   & 2020                   & {\color[HTML]{000000} 21.64}                                        & {\color[HTML]{000000} 17.16}                & {\color[HTML]{000000} 29.70}                                        & 27.99 & 34.31 & \cellcolor[HTML]{EFEFEF}28.32                                       & 26.39 & 32.47 & \cellcolor[HTML]{EFEFEF}26.69                                       & 21.59                         & 24.73                         & 21.02                                                               & {\color[HTML]{3531FF} { \textbf{41.96}}}     & {\color[HTML]{000000} 11.21}  & {\color[HTML]{000000} 12.89}  & {\color[HTML]{3531FF} { \textbf{10.96}}} & {\color[HTML]{3531FF} {\textbf{15.0}}}                                           \\ \hline
{$w_{2020}$}                   & 2021                   & \cellcolor[HTML]{EFEFEF}{\color[HTML]{3531FF} { \textbf{24.02}}} & {\color[HTML]{3531FF} { \textbf{18.32}}} & \cellcolor[HTML]{EFEFEF}{\color[HTML]{3531FF} { \textbf{31.71}}} & 31.59 & 35.98 & \cellcolor[HTML]{EFEFEF}{\color[HTML]{3531FF} { \textbf{31.71}}} & 29.69 & 33.83 & \cellcolor[HTML]{EFEFEF}{\color[HTML]{3531FF} { \textbf{29.79}}} & \cellcolor[HTML]{DAE8FC}22.20 & \cellcolor[HTML]{DAE8FC}24.83 & \cellcolor[HTML]{DAE8FC}{\color[HTML]{3531FF} { \textbf{21.57}}} & \cellcolor[HTML]{EFEFEF}45.06                   & \cellcolor[HTML]{DAE8FC}10.57 & \cellcolor[HTML]{DAE8FC}12.03 & \cellcolor[HTML]{DAE8FC}10.32               & 16.0                                           \\ \hline
\end{tabular}
}
\caption{\label{tab_2}\small\textbf{GPT2 Medium}: Transfer learning model comparison across yearly FA data (year-$2020$ $\in \mathbb{R}^{5672\times 79}$; year-$2021$ $\in \mathbb{R}^{2906\times 79}$). The checkpoint weights from previous year (for instance, year 2019 is used for training year -$2020$ and so on). The marginal improvement in the LESE-3 scores for year-$2021$ suggests that the weight ($w_{2019}$), generalizes the sub-domains. Improvement in LESE-3 is not monotonic, observe the best scores (in {\color{blue}{\textbf{bold-blue}}}) for year-$2020$ and only slightly less for year-$2021$ due to insufficient learning data. $w^{\dagger}$: Fine-tuned pre-trained weight.}
\end{table} 

The best performing model on longest common triplets (LCT), related to longest common subsequence generation is GPT2 Medium and Large. Both ROUGE-L and METEOR favor long sequence triplet generation. Similarly, GPT2 Medium, measures up in performance with GPT2 Large (in terms of ROUGE-L and METEOR). In the next section, where we compare the model FATG to expert (human) triplets, we show that the FATs generated by GPT2 Medium and GPT2 Large are not differentiable.
\subsection{Qualitative Evaluation: FATG}
The quality of FATG of a model depends on its ability to adapt previously seen source domain knowledge to downstream domain task (at training time) as well as the quality of prompt (at generation time). Well preprocessed training data ($\Lambda$) and prompt (failure description), $x$ are two of the major factors that can degrade the generative quality. We classify this generation difficulty into three classes \texttt{(i)} Short-sequence FATG \texttt{(ii)} Long-sequence FATG. Transformers-based PLMs are by default designed for long sequence generation and particularly perform well for generating contextual text and summarization task. We evaluate their performance for structured FATG and report our findings.
\\
\subsubsection{Short sequence FATG (SS-FATG)}
\begin{table}[!b]
\centering
\setlength{\tabcolsep}{0.5em} 
{\renewcommand{\arraystretch}{1.7}
\tiny
\begin{tabular}{llll}
\hline
\texttt{FDR}                & \multicolumn{2}{l}{{\color[HTML]{FE0000} \texttt{failed abnormal thd analysis failure manufacturing axis customer limit complaint castelletto data}}}                                                                                                                                                                                                                                                                                                                                                                                                                                                                                                                                                                                                                                                                            \\ \hline
                   & \texttt{FATs}                                                                                                                                                                                                                                                                                                                                                                                                                                                                                                                                                                                                                                                                                                                                                        & \cellcolor[HTML]{DAE8FC}\texttt{LESE-1 (\%)} & \cellcolor[HTML]{DAE8FC}\texttt{LESE-3 (\%)} \\ \hline
FATs (Human) & \begin{tabular}[c]{@{}l@{}}$\blacktriangleright$ Others; administrative activity; OM113 - LEICA M165C \\$\blacktriangleright$ Non destructive Inspection; X-ray; PK103 - PHOENIX X-RAY NANOMEX \\$\blacktriangleright$ Electrical Failure Verification; Electrical parametrical test; ZZ003 - CRI7 \\$\blacktriangleright$ Electrical Failure Verification; Electrical parametrical test; ZZ002 - CRI6\end{tabular}                                                                                                                                                                                                                                                                                                                                                                                                                              & \cellcolor[HTML]{DAE8FC}100 & \cellcolor[HTML]{DAE8FC}100        \\ \hline
mini-GPT           &      -                                                                                                                                                                                                                                                                                                                                                                                                                                                                                                                                                                                                                                                                                                                                                       & \cellcolor[HTML]{DAE8FC} - & \cellcolor[HTML]{DAE8FC}  -       \\ \hline
BART            & \begin{tabular}[c]{@{}l@{}}$\blacktriangleright$ d abnormal thd \\$\blacktriangleright$ Physical Analysis; Optical inspection; ANALYSIS EQU EQUIPMENT \\$\blacktriangleright$ Others; administrative failed; thD403 \\$\blacktriangleright$ Others; administrative activity; FINAL REPORT GENERATION \\$\blacktriangleright$ Non destructive Inspection; Optical microscopy; LOW POWER SCOPE\end{tabular}                                                                                                                                                                                                                                                                                                                                                                                                                                                                     & \cellcolor[HTML]{DAE8FC}0.00   & \cellcolor[HTML]{DAE8FC}0.00  \\ \hline
GPT3        & \begin{tabular}[c]{@{}l@{}}$\blacktriangleright$ sample preparation package; decap; tools \\$\blacktriangleright$ sample preparation; package decap; sp001 - wet + microscopio \\$\blacktriangleright$ physical analysis; sem; em202 - nvision 40 \\$\blacktriangleright$ others; administrative activity; om103 - leica dm lmp \\$\blacktriangleright$ others; administrative activity; om113 - leica m165c \\$\blacktriangleright$ non destructive inspection; ir microscopy; om111 - olympus mx51 - f \\$\blacktriangleright$ electrical failure verification; electrical parametrical test; zz003 - cri7 ( m202 \\$\blacktriangleright$ global fault localisation; emmi; failure rate chain chain analyzer ( ecn \\$\blacktriangleright$ global fault localisation; emmi; fi105 - s105 ) oem esafm01 \\$\blacktriangleright$ electrical failure verification; electrical parametrical test; fv001 - bench test \\$\blacktriangleright$ electrical failure verification; continuity test; ftracer - ecn 4\end{tabular} & \cellcolor[HTML]{DAE8FC}35.56 & \cellcolor[HTML]{DAE8FC}13.33      \\ \hline
GPT2            & {\color[HTML]{333333} \begin{tabular}[c]{@{}l@{}}$\blacktriangleright$ Others; administrative activity; OM113 - LEICA M165C \\$\blacktriangleright$ Non destructive Inspection; X-ray; PK103 - PHOENIX X-RAY NANOMEX \\$\blacktriangleright$ Electrical Failure Verification; Electrical parametrical test; FV001 - BENCH TEST\end{tabular}}                                                                                                                                                                                                                                                                                                                                                                                                                                                                                & \cellcolor[HTML]{DAE8FC} 76.19 & \cellcolor[HTML]{DAE8FC} 57.14           \\ \hline
GPT2 Medium     & \begin{tabular}[c]{@{}l@{}}$\blacktriangleright$ Others; administrative activity; OM113 - LEICA M165C \\$\blacktriangleright$ Non destructive Inspection; X-ray; PK103 - PHOENIX X-RAY NANOMEX \\$\blacktriangleright$ Electrical Failure Verification; Electrical parametrical test; ZZ003 - CRI7 \\$\blacktriangleright$ Electrical Failure Verification; Electrical parametrical test; ZZ002 - CRI6\end{tabular}                                                                                                                                                                                                                                                                                                                                                                                                                              & \cellcolor[HTML]{DAE8FC}\color[HTML]{3531FF}\textbf{100} & \cellcolor[HTML]{DAE8FC}\color[HTML]{3531FF}\textbf{100}        \\ \hline
GPT2 Large      & \begin{tabular}[c]{@{}l@{}}$\blacktriangleright$ Others; administrative activity; OM113 - LEICA M165C \\$\blacktriangleright$ Non destructive Inspection; X-ray; PK103 - PHOENIX X-RAY NANOMEX \\$\blacktriangleright$ Electrical Failure Verification; Electrical parametrical test; FV001 - BENCH TEST\end{tabular}                                                                                                                                                                                                                                                                                                                                                                                                                                                                                                             & \cellcolor[HTML]{DAE8FC} 76.19 & \cellcolor[HTML]{DAE8FC} 57.14           \\ \hline
\end{tabular}
\caption{\label{ssfatg}\small \textbf{SS-FATG}. A comparison of the failure analysis generated for different models given a prompt with $4$-human expert FATs. GPT2-Medium performs the best in generating highly correlating FATs as the human expert with a perfect LESE-$1$ and LESE-$3$ scores.}
}
\end{table}

\begin{table}[!b]
\centering
\setlength{\tabcolsep}{0.5em} 
{\renewcommand{\arraystretch}{1.7}
\tiny
\renewcommand{\tabcolsep}{2.5pt}
\begin{tabular}{llll}
\hline
FDR                & \multicolumn{2}{l}{{\color[HTML]{FE0000} \texttt{fail fault support serial number isolation analysis failure lotid shenzhen engineering postda determine}}}                                                                                                                                                                                                                                                                                                                                                                                                                                                                                                                                                                 \\ \hline
                   & FATs                                                                                                                                                                                                                                                                                                                                                                                                                                                                                                                                                                                                                                                                          & \cellcolor[HTML]{DAE8FC}\texttt{LESE-1} (\%) & \cellcolor[HTML]{DAE8FC}\texttt{LESE-3} (\%) \\ \hline
FATs (Human) & \begin{tabular}[c]{@{}l@{}}$\blacktriangleright$ Sample preparation; Wafer / Die Backside prep; SP120 - ALLIED TECHPREP \\$\blacktriangleright$ Sample preparation; Die Delayering; SP001 - WET+MICROSCOPIO \\$\blacktriangleright$ Non destructive Inspection; SEM; EM201 - LEO 1530 \\$\blacktriangleright$ Global fault localisation; Static Laser Techniques; FI105 - IPHEMOS-MP \\$\blacktriangleright$ Electrical Failure Verification; Continuity test; FV001 - BENCH TEST\end{tabular}                                                                                                                                                                                                                                                                                                    & \cellcolor[HTML]{DAE8FC}100  & \cellcolor[HTML]{DAE8FC}100      \\ \hline
mini-GPT           &      -                                                                                                                                                                                                                                                                                                                                                                                                                                                                                                                                                                                                                                                                         & \cellcolor[HTML]{DAE8FC}-  & \cellcolor[HTML]{DAE8FC}  -       \\ \hline
BART            & \begin{tabular}[c]{@{}l@{}}$\blacktriangleright$ d abnormal thd anomaly failed thc abnormal parametric gnd amp \\$\blacktriangleright$ Sample preparation Package decap SP006 - MAN\_LASER+WET+MICRO \\$\blacktriangleright$ Non destructive Inspection X-ray PK103 - PHOENIX X-RAY\end{tabular}                                                                                                                                                                                                                                                                                                                                                                                                                                              & \cellcolor[HTML]{DAE8FC}16.67 & \cellcolor[HTML]{DAE8FC}0.00      \\ \hline
GPT3        & \begin{tabular}[c]{@{}l@{}}$\blacktriangleright$ sample preparation package decap tools \\$\blacktriangleright$ sample preparation; package decap; sp001 - wet + microscopio \\$\blacktriangleright$ physical analysis; sem; em202 - nvision 40 \\$\blacktriangleright$ others; administrative activity; om103 - leica dm lmp \\$\blacktriangleright$ others; administrative activity; om113 - leica m165c \\$\blacktriangleright$ non destructive inspection; ir microscopy; om111 - olympus mx51 - f \\$\blacktriangleright$ electrical failure verification; electrical parametrical test; zz003 - cri7 \\$\blacktriangleright$ global fault localisation; static laser techniques; fi105 - iphemos \\$\blacktriangleright$ global fault localisation; emmi; emmi obirch ch \\$\blacktriangleright$ electrical failure verification; electrical parametrical test; fv001 - bench test\end{tabular} & \cellcolor[HTML]{DAE8FC}0.00   & \cellcolor[HTML]{DAE8FC}0.00   \\ \hline
GPT2            & {\color[HTML]{333333} \begin{tabular}[c]{@{}l@{}}$\blacktriangleright$ rejects shz obirch pocutornplgattransport \\$\blacktriangleright$ Sample preparation; Die Delayering; PARALLEL LAPPING-ECN6432 \\$\blacktriangleright$ Physical Analysis; SEM; FIB/SEM \\$\blacktriangleright$ Physical Analysis; FIB Cross Section; FIB/SEM \\$\blacktriangleright$ Non destructive Inspection; Optical microscopy; METALGRAPHIC MICROSCOPE MX61 \\$\blacktriangleright$ Global fault localisation; Static Laser Techniques; EMMI OBIRCH \\$\blacktriangleright$ Global fault localisation; EMMI; EMMI OBIRCH \\$\blacktriangleright$ Electrical Failure Verification; Electrical parametrical test; BENCH TEST\end{tabular}}                                                                                                                                  & \cellcolor[HTML]{DAE8FC}30.77 & \cellcolor[HTML]{DAE8FC}0.00     \\ \hline
GPT2 Medium     & \begin{tabular}[c]{@{}l@{}}$\blacktriangleright$ Sample preparation; Package decap; TOOL DECAP \\$\blacktriangleright$ Physical Analysis; SEM; FIB/SEM \\$\blacktriangleright$ Physical Analysis; FIB Cross Section; FIB/SEM \\$\blacktriangleright$ Non destructive Inspection; Optical microscopy; METALGRAPHIC MICROSCOPE MX61 \\$\blacktriangleright$ Global fault localisation; Static Laser Techniques; EMMI OBIRCH \\$\blacktriangleright$ Electrical Failure Verification; Continuity test; CURVE TRACER 1\end{tabular}                                                                                                                                                                                                                                                                              & \cellcolor[HTML]{DAE8FC}\color[HTML]{3531FF}\textbf{36.36} & \cellcolor[HTML]{DAE8FC}0.00      \\ \hline
GPT2 Large      & \begin{tabular}[c]{@{}l@{}}$\blacktriangleright$ Sample preparation; Package decap; TOOL DECAP \\$\blacktriangleright$ Physical Analysis; SEM; FIB/SEM \\$\blacktriangleright$ Physical Analysis; FIB Cross Section; FIB/SEM \\$\blacktriangleright$ Non destructive Inspection; Optical microscopy; METALGRAPHIC MICROSCOPE MX61 \\$\blacktriangleright$ Global fault localisation; Static Laser Techniques; EMMI OBIRCH \\$\blacktriangleright$ Global fault localisation; EMMI; EMMI OBIRCH \\$\blacktriangleright$ Electrical Failure Verification; Electrical parametrical test; BENCH TEST\end{tabular}                                                                                                                                                                                                                     & \cellcolor[HTML]{DAE8FC}27.78  & \cellcolor[HTML]{DAE8FC}0.00    \\ \hline
\end{tabular}
\caption{\label{ssfatg_2}\small \textbf{SS-FATG}. A comparison of the failure analysis generated for different models given a prompt with $5$-human expert FATs. LESE-3 scores for this FDR is null indicating the absence of correlating triplets with human FATs. GPT-2 Medium LESE-1 scores outperform other models signaling quality generation of step types and substep techniques.}
}
\end{table}
\begin{table}[!b]
\centering
\setlength{\tabcolsep}{0.5em} 
{\renewcommand{\arraystretch}{1.7}
\tiny
\begin{tabular}{llll}
\hline
FDR                & \multicolumn{2}{l}{{\color[HTML]{FE0000} \begin{tabular}[c]{@{}l@{}}\texttt{result serial pras visual assembly coppell cse stated field ale preanalysis michael incident}\\ \texttt{passivation dodge comm stmicroelectronics required material eos scott customer} \\ \texttt{segment production belvidere conti ppm final failure registration hole supplier} \\ \texttt{complaint chr relevant cseale shorted severirty analyst product standard}\\ \texttt{analysis ecu type location inspection vin mileage quantity warranty subtype}\end{tabular}}}                                                                                          \\ \hline
                   & FATs                                                                                                                                                                                                                                                                                                                                                                                                                                                                                                                                                              & \cellcolor[HTML]{DAE8FC}\texttt{LESE-1} (\%) & \cellcolor[HTML]{DAE8FC}\texttt{LESE-3} (\%) \\ \hline
FATs (Human) & \begin{tabular}[c]{@{}l@{}}$\blacktriangleright$ Physical Analysis; Optical inspection; ANALYSIS EQUIPMENT \\$\blacktriangleright$ Physical Analysis; Optical inspection; HIGH POWER SCOPE \\$\blacktriangleright$ Physical Analysis; Optical inspection; LOW POWER SCOPE\end{tabular}                                                                                                                                                                                                                                                                                                                                                  & \cellcolor[HTML]{DAE8FC}100  & \cellcolor[HTML]{DAE8FC}100       \\ \hline
mini-GPT           &        -                                                                                                                                                                                                                                                                                                                                                                                                                                                                                                                                                           & \cellcolor[HTML]{DAE8FC} - & \cellcolor[HTML]{DAE8FC}   -   \\ \hline
BART            & \begin{tabular}[c]{@{}l@{}}$\blacktriangleright$ d abnormal thd \\$\blacktriangleright$ Physical Analysis; Optical inspection; ANALYSISIS EQUIPMENT \\$\blacktriangleright$ Others; administrative; failed tht abnormal observed mud abnormal \\$\blacktriangleright$ Inspection; Optical inspection; HIGH POWER failed STd abnormal \\$\blacktriangleright$ Non destructive Inspection; Optical microscopy; LOW POWER SCOPE\end{tabular}                                                                                                                                                                                                                                           & \cellcolor[HTML]{DAE8FC}19.05 & \cellcolor[HTML]{DAE8FC}0.00      \\ \hline
GPT3        & \begin{tabular}[c]{@{}l@{}}$\blacktriangleright$ sample preparation; package decap; tools \\$\blacktriangleright$ sample preparation; package decap; sp001 - wet + microscopio \\$\blacktriangleright$ physical analysis; sem; em202 - nvision 40 \\$\blacktriangleright$ others; administrative activity; om103 - leica dm lmp \\$\blacktriangleright$ non destructive inspection; ir microscopy; om111 - olympus mx51 - f \\$\blacktriangleright$ electrical failure verification; electrical parametrical test; fv001 - bench test \\$\blacktriangleright$ global fault localisation; emmi; fibsem \\$\blacktriangleright$ global fault localisation; emmi; analyzer \\$\blacktriangleright$ global fault localisation; emmi; lsm \\
electrical failure verification\end{tabular} & \cellcolor[HTML]{DAE8FC}5.56 & \cellcolor[HTML]{DAE8FC}0.00     \\ \hline
GPT2            & {\color[HTML]{333333} \begin{tabular}[c]{@{}l@{}}$\blacktriangleright$ signature andorch conditions huawei \\$\blacktriangleright$ Physical Analysis; Optical inspection; LSM \\$\blacktriangleright$ Physical Analysis; Optical inspection; L200 MICROSCOPE 2 \\$\blacktriangleright$ Others; administrative activity; FINAL REPORT GENERATION \\$\blacktriangleright$ Global fault localisation; Static Laser Techniques; LSM\end{tabular}}                                                                                                                                                                                                                               & \cellcolor[HTML]{DAE8FC}\color[HTML]{3531FF}\textbf{34.78} & \cellcolor[HTML]{DAE8FC}0.00      \\ \hline
GPT2 Medium     & \begin{tabular}[c]{@{}l@{}}$\blacktriangleright$ conti severirty preanalysis final coppell \\$\blacktriangleright$ Sample preparation; Package decap; CHEMICAL HOOD \\$\blacktriangleright$ Sample preparation; Package decap; LASER DECAPPER \\$\blacktriangleright$ Physical Analysis; Optical inspection; L200 MICROSCOPE 2 \\$\blacktriangleright$ Physical Analysis; Optical inspection; NIKON STEREO MICROSCOPE IN CPL FA \\$\blacktriangleright$ Others; administrative activity; FINAL REPORT GENERATION\end{tabular}                                                                                                                                                                    & \cellcolor[HTML]{DAE8FC}30.77    & \cellcolor[HTML]{DAE8FC}0.00  \\ \hline
GPT2 Large      & \begin{tabular}[c]{@{}l@{}}$\blacktriangleright$ conti cseale \\$\blacktriangleright$ Sample preparation; Package decap; CHEMICAL HOOD \\$\blacktriangleright$ Sample preparation; Package decap; LASER DECAPPER \\$\blacktriangleright$ Physical Analysis; Optical inspection; L200 MICROSCOPE 2 \\$\blacktriangleright$ Physical Analysis; Optical inspection; NIKON STEREO MICROSCOPE IN CPL FA \\$\blacktriangleright$ Others; administrative activity; FINAL REPORT GENERATION\end{tabular}                                                                                                                                                                                                           & \cellcolor[HTML]{DAE8FC}33.33   & \cellcolor[HTML]{DAE8FC}0.00   \\ \hline
\end{tabular}
\caption{\label{ssfatg_3}\small \textbf{SS-FATG}. A comparison of the failure analysis generated for different models given a prompt with $3$-human expert FATs. GPT2 performs the best in generating high correlating FATs as the human expert.}
}
\end{table}

Given a FDR prompt and a short human-expert failure analysis triplets (with length ranging between $3-6$), Short sequence FATG (\textbf{SS-FATG}) is the ability of a fine-tuned PLM to generate the same order and length of the FATs given by human-expert. Short sequence FATs (\textbf{SS-FATs}), can be partial incomplete FA that compliments a main FA. We observe from generation task, that all the models including mini-GPT (non-pretrained baseline), BART, GPT2 (base), GPT2 Medium and GPT2 Large are capable of generating short sequences. However, PT-BART generates incorrect SS-FATs while GPT3 generates mostly long sequence with low BLEU and LESE scores. GPT3 is modeled with the objective of generating long sequences for story-telling, hence, its behaviour during FATG. We give an illustration of a case of SS-FATG task using the best result from GPT2 and a much lesser BART generation result.\\
\small
\colorbox{red!10}{Short sequence FATG (SS-FATG)}\\
\colorbox{lightgray!25}{FDR prompt:} \texttt{failed abnormal thd analysis failure manufacturing axis customer limit complaint castelletto data}\\
\colorbox{lightgray!25}{Human FATs}:\\
$\blacktriangleright$ \texttt{Others; administrative activity; OM113 - LEICA M165C}\\
$\blacktriangleright$ \texttt{Non destructive Inspection; X-ray; PK103 - PHOENIX X-RAY NANOMEX}\\
$\blacktriangleright$ \texttt{Electrical Failure Verification; Electrical parametrical test; ZZ003 - CRI7}\\
$\blacktriangleright$ \texttt{Electrical Failure Verification; Electrical parametrical test; ZZ002 - CRI6}\\
\colorbox{lightgray!25}{GPT2-M FATs:} \colorbox{blue!7}{BLEU-3: $100.00\%$} \colorbox{cyan!14}{LESE-3: Prec.: $100.00\%$ Rec.: $100.00\%$ F1.: $100.00\%$}\\
$\blacktriangleright$ \texttt{\colorbox{green!10}{Others}; \colorbox{green!10}{administrative activity}; \colorbox{green!10}{OM113 - LEICA M165C}} \\
$\blacktriangleright$ \texttt{\colorbox{green!10}{Non destructive Inspection}; \colorbox{green!10}{X-ray}; \colorbox{green!10}{PK103 - PHOENIX X-RAY NANOMEX}} \\
$\blacktriangleright$ \texttt{\colorbox{green!10}{Electrical Failure Verification}; \colorbox{green!10}{Electrical parametrical test}; \colorbox{green!10}{ZZ00 3 - CRI7}} \\
$\blacktriangleright$ \texttt{\colorbox{green!10}{Electrical Failure Verification}; \colorbox{green!10}{Electrical parametrical test}; \colorbox{green!10}{ZZ002 - CRI6}} \\
\colorbox{lightgray!25}{BART FATs:} \colorbox{blue!7}{BLEU-3: $0.00\%$} \colorbox{cyan!14}{LESE-3: Prec.: $0.00\%$ Rec.: $0.00\%$ F1.: $0.00\%$}\\
$\blacktriangleright$ \texttt{\colorbox{lightgray!65}{d abnormal thd}}\\
$\blacktriangleright$\texttt{\colorbox{lightgray!65}{Physical Analysis}; \colorbox{lightgray!65}{Optical inspection}; \colorbox{lightgray!65}{ANALYSIS EQU EQUIPMENT}}\\
$\blacktriangleright$\texttt{\colorbox{lightgray!65}{Others}; \colorbox{lightgray!65}{administrative}; \colorbox{lightgray!65}{failed thD403}}\\
$\blacktriangleright$\texttt{\colorbox{green!10}{Others}; \colorbox{green!10}{administrative activity}; \colorbox{lightgray!65}{FINAL REPORT GENERATION}}\\
$\blacktriangleright$\texttt{\colorbox{green!10}{Non destructive Inspection}; \colorbox{lightgray!65}{Optical microscopy}; \colorbox{lightgray!65}{LOW POWER SCOPE}}
\normalsize

The light green highlights are the correct FA generations while the grey highlight are incorrect FA generation. We observe that all FATs generated by the fine-tuned GPT2 Medium are correct and ordered compared to those generated by fine-tuned BART model. A complete table containing generation of other models for the same prompt is given below (see Table \ref{ssfatg}). Other SS-FATG examples are given in Tables (\ref{ssfatg_2} and \ref{ssfatg_3}). 
\subsubsection{Long sequence FATG}.
Long sequence FATG (\textbf{LS-FATG}) is the ability of a fine-tuned PLM to generate long-length sequences that preserve the sequential order as the original human-expert FATs. Long sequence FATs (\textbf{LS-FATs}) are complete failure analysis containing all sequence of FAs in the correct order. They are usually above $7$ or more triplets, depending on the number of locations the failure device is transferred before finding the root cause. Attention mechanism in transformers are modeled to generate long text sequences, for instance, in the case of DialogGPT \citep{dialog_gpt} for conversational response generation, as well as for text summarization. In this context of FATG, their performance score is attenuated since we are modeling structured data.\\
The best LESE-1 score obtained for a FDR prompt having $7$-FATs is $31\%$ with fine-tuned GPT2 (Base and Large). Despite been suited for long text generation, GPT3 fails to generate triplets consistent with human FATs. Nothing-withstanding, GPT3 generates FATs consistent with what an expert would propose for similar failure description. In generating LS-FATs, none of the models accurately generated the correct equipment proposed by human-expert, however, the equipment generated could serve similar purpose as that proposed by expert.\\
\begin{table}[!ht] 
\centering
\setlength{\tabcolsep}{0.5em} 
{\renewcommand{\arraystretch}{1.7}
\tiny
\renewcommand{\tabcolsep}{2.5pt}
\begin{tabular}{llll}
\hline
FDR                & \multicolumn{2}{l}{{\color[HTML]{FE0000} \texttt{failure customer complaint amp reliability analysis shenzhen}}}                                                                                                                                                                                                                                                                                                                                                                                                                                                                                                                          \\ \hline
                   & \texttt{FATs}                                                                                                                                                                                                                                                                                                                                                                                                                                                                                                                                                                                        & \cellcolor[HTML]{DAE8FC}\texttt{LESE-1} (\%) & \cellcolor[HTML]{DAE8FC}\texttt{LESE-3} (\%) \\ \hline
FATs (Human) & \begin{tabular}[c]{@{}l@{}}$\blacktriangleright$ Sample preparation; Package decap; MANUAL CHEMICAL DECAP \\$\blacktriangleright$ Physical Analysis; SEM; PHILIPS SEM XL40 (RSE001) \\$\blacktriangleright$ Non destructive Inspection; X-ray; B800\_FA-XR01 \\$\blacktriangleright$ Physical Analysis; SEM; HITACHI SEM TM-1000 (RSE002) \\$\blacktriangleright$ Non destructive Inspection; Optical microscopy; OLYMPUS MICROSCOPE SZ40 (ROS005) \\$\blacktriangleright$ Electrical Failure Verification; Continuity test; TEKTRONIX 370A CURVE TRACER \\$\blacktriangleright$ Electrical Failure Verification; Continuity test; KEITHLEY PARAMETRIC ANALYZER\end{tabular}                                                                    & \cellcolor[HTML]{DAE8FC}100   & \cellcolor[HTML]{DAE8FC}100     \\ \hline
mini-GPT           &           -                                                                                                                                                                                                                                                                                                                                                                                                                                                                                                                                                                                  & \cellcolor[HTML]{DAE8FC}-  & \cellcolor[HTML]{DAE8FC}-         \\ \hline
BART            & \begin{tabular}[c]{@{}l@{}}$\blacktriangleright$ d abnormal thd abnormalities failed tht abnormal parametric gnd amp \\$\blacktriangleright$ Physical Analysis; FIB Cross Section; EM202 - NVISION 40 \\$\blacktriangleright$ Others Others; administrative activity; OM800 - LEICA - FA \\$\blacktriangleright$ Others; administrative; failed THdd abnormal\end{tabular}                                                                                                                                                                                                                                                                                                       & \cellcolor[HTML]{DAE8FC}6.25   & \cellcolor[HTML]{DAE8FC}0.00   \\ \hline
GPT3        & \begin{tabular}[c]{@{}l@{}}$\blacktriangleright$ sample preparation; package decap; tools \\$\blacktriangleright$ sample preparation; package decap; sp001 - wet + microscopio \\$\blacktriangleright$ physical analysis; sem; em202 - nvision 40 \\$\blacktriangleright$ others; administrative activity; om103 - leica dm lmp \\$\blacktriangleright$ non destructive inspection; ir microscopy; om111 - olympus mx51 - f \\$\blacktriangleright$ electrical failure verification; electrical parametrical test; fv001 - bench test \\$\blacktriangleright$ global fault localisation; sample preparation; dies localisation tracernernbased fault \\$\blacktriangleright$ electrical failure verification; dies marginal parametrical test; 401 - fa\end{tabular} & \cellcolor[HTML]{DAE8FC}22.22  & \cellcolor[HTML]{DAE8FC}0.00    \\ \hline
GPT2            & {\color[HTML]{333333} \begin{tabular}[c]{@{}l@{}}$\blacktriangleright$ Sample preparation; Package decap; CHEMICALDECAP ACID \\$\blacktriangleright$ Sample preparation; Package decap; HOTPLATE \\$\blacktriangleright$ Physical Analysis; Optical inspection; MICROSCOPE MX51 \\$\blacktriangleright$ Physical Analysis; Optical inspection; METALGRAPHIC MICROSCOPE MX61 \\$\blacktriangleright$ Non destructive Inspection; Optical microscopy; MICROSCOPE VH8000 \\$\blacktriangleright$ Non destructive Inspection; X-ray; 3D X-RAY \\$\blacktriangleright$ Electrical Failure Verification; Continuity test; CURVE TRACER 2 \\$\blacktriangleright$ Non destructive Inspection; Optical microscopy; MICROSCOPE MX51\end{tabular}}                             & \cellcolor[HTML]{DAE8FC}\color[HTML]{3531FF}\textbf{31.11} & \cellcolor[HTML]{DAE8FC}0.00      \\ \hline
GPT2 Medium     & \begin{tabular}[c]{@{}l@{}}$\blacktriangleright$ Sample preparation; Package decap; CHEMICALDECAP ACID \\$\blacktriangleright$ Sample preparation; Package decap; HOTPLATE \\$\blacktriangleright$ Physical Analysis; Optical inspection; MICROSCOPE MX51 \\$\blacktriangleright$ Physical Analysis; Optical inspection; METALGRAPHIC MICROSCOPE MX61 \\$\blacktriangleright$ Non destructive Inspection; Optical microscopy; MICROSCOPE VH8000 \\$\blacktriangleright$ Non destructive Inspection; X-ray; 3D X-RAY \\$\blacktriangleright$ Electrical Failure Verification; Continuity test; CURVE TRACER 2 \\$\blacktriangleright$ Non destructive Inspection; Optical microscopy; MICROSCOPE MX51\end{tabular}                                                    & \cellcolor[HTML]{DAE8FC}29.79   & \cellcolor[HTML]{DAE8FC}0.00   \\ \hline
GPT2 Large      & \begin{tabular}[c]{@{}l@{}}$\blacktriangleright$ Sample preparation; Package decap; CHEMICALDECAP ACID \\$\blacktriangleright$ Sample preparation; Package decap; HOTPLATE \\$\blacktriangleright$ Physical Analysis; Optical inspection; MICROSCOPE MX51 \\$\blacktriangleright$ Physical Analysis; Optical inspection; METALGRAPHIC MICROSCOPE MX61 \\$\blacktriangleright$ Non destructive Inspection; Optical microscopy; MICROSCOPE VH8000 \\$\blacktriangleright$ Non destructive Inspection; X-ray; 3D X-RAY \\$\blacktriangleright$ Electrical Failure Verification; Continuity test; CURVE TRACER 2 \\$\blacktriangleright$ Non destructive Inspection; Optical microscopy; MICROSCOPE MX51\end{tabular}                                                    & \cellcolor[HTML]{DAE8FC}\color[HTML]{3531FF}\textbf{31.11}  & \cellcolor[HTML]{DAE8FC}0.00    \\ \hline
\end{tabular}
\caption{\small\label{ls_fatg_1} \textbf{LS-FATG}.Failure analysis triplets generated for the different models. LESE-1 scores for GPT2-Large and Base are best performing in generating correlating FATs as the human expert.}
}
\end{table}\\ 
\small
\colorbox{red!10}{Long sequence FATG (LS-FATG)}\\
\colorbox{lightgray!25}{FDR prompt:} \texttt{failure customer complaint amp reliability analysis shenzhen}\\
\colorbox{lightgray!25}{Human FATs:}\\
$\blacktriangleright$ \texttt{Sample preparation; Package decap; MANUAL CHEMICAL DECAP}\\
$\blacktriangleright$ \texttt{Physical Analysis; SEM; PHILIPS SEM XL40 (RSE001)}\\
$\blacktriangleright$ \texttt{Non destructive Inspection; X-ray; B800 FA-XR01}\\
$\blacktriangleright$ \texttt{Physical Analysis; SEM; HITACHI SEM TM-1000 (RSE002)}\\
$\blacktriangleright$ \texttt{Non destructive Inspection; Optical microscopy; OLYMPUS MICROSCOPE SZ40 (ROS005)}\\
$\blacktriangleright$ \texttt{Electrical Failure Verification; Continuity test; TEKTRONIX 370A CURVE TRACER}\\
$\blacktriangleright$ \texttt{Electrical Failure Verification; Continuity test; KEITHLEY PARAMETRIC ANALYZER}\\
\colorbox{lightgray!25}{GPT2-M FATs:} \colorbox{blue!7}{BLEU-3: $0.00\%$} \colorbox{cyan!14}{LESE-3: Prec.: $0.00\%$ Rec.: $0.00\%$ F1.: $0.00\%$}\\
$\blacktriangleright$ \texttt{\colorbox{green!10}{Sample preparation}; \colorbox{green!10}{Package decap}; \colorbox{lightgray!65}{CHEMICALDECAP ACID}} \\
$\blacktriangleright$ \texttt{\colorbox{lightgray!65}{Sample preparation}; \colorbox{lightgray!65}{Package decap}; \colorbox{lightgray!65}{HOTPLATE}} \\
$\blacktriangleright$ \texttt{\colorbox{green!10}{Physical Analysis}; \colorbox{lightgray!65}{Optical inspection}; \colorbox{lightgray!65}{MICROSCOPE MX51}} \\
$\blacktriangleright$ \colorbox{lightgray!65}{Physical Analysis}; \colorbox{lightgray!65}{Optical inspection}; \colorbox{lightgray!65}{METALGRAPHIC MICROSCOPE MX61} \\
$\blacktriangleright$ \texttt{\colorbox{green!10}{Non destructive Inspection}; \colorbox{green!10}{Optical microscopy}; \colorbox{lightgray!65}{MICROSCOPE VH8000}} \\
$\blacktriangleright$ \texttt{\colorbox{green!10}{Non destructive Inspection}; \colorbox{green!10}{X-ray}; \colorbox{lightgray!65}{3D X-RAY}} \\
$\blacktriangleright$ \texttt{\colorbox{green!10}{Electrical Failure Verification}; \colorbox{green!10}{Continuity test}; \colorbox{lightgray!65}{CURVE TRACER 2}} \\
$\blacktriangleright$ \texttt{\colorbox{lightgray!65}{Non destructive Inspection}; \colorbox{lightgray!65}{Optical microscopy}; \colorbox{lightgray!65}{MICROSCOPE MX51}} \\
\colorbox{lightgray!25}{BART FATs:} \colorbox{blue!7}{BLEU-3: $0.00\%$} \colorbox{cyan!14}{LESE-3: Prec.: $0.00\%$ Rec.: $0.00\%$ F1.: $0.00\%$}\\
$\blacktriangleright$ \texttt{\colorbox{lightgray!65}{d abnormal thd abnormalities failed tht abnormal \colorbox{green!10}{parametric} gnd amp}}\\
$\blacktriangleright$\texttt{\colorbox{green!10}{Physical Analysis}; \colorbox{lightgray!65}{FIB Cross Section}; \colorbox{lightgray!65}{EM202 - NVISION 40}}\\
$\blacktriangleright$\texttt{\colorbox{lightgray!65}{Others Others}; \colorbox{lightgray!65}{administrative activity}; \colorbox{lightgray!65}{OM800 - LEICA - FA}}\\
$\blacktriangleright$\texttt{\colorbox{lightgray!65}{Others}; \colorbox{lightgray!65}{administrative}; \colorbox{lightgray!65}{THdd abnormal}}
\normalsize

Again, we observe all GPT2 models perform on par for generating long sequences when compared with BART and GPT3 (See full comparison on Table \ref{ls_fatg_1}). It is very clear that the order of performance of Transformer for structured SS-FATG task is $\texttt{GTP2 M \& L} > \texttt{GPT2} > \texttt{BART} > \texttt{GPT3}$ and $\texttt{GTP2 M \& L} > \texttt{GPT2} > \texttt{GPT3} > \texttt{BART}$ for LS-FATG.
Furthermore, depending on the description and location of the FA, different FATs can be generated with keywords of the location of the FA in response to the prompt. This is illustrated in the example below,\\
\small
\colorbox{red!10}{Long sequence FATG (LS-FATG)}\\
\colorbox{lightgray!25}{FDR prompt:} \texttt{GNB-PQE\_18\_00950 BALAN ECC51 O1T short AVDD 1 part received O1T with short AVDD lot no traceability}\\
\colorbox{lightgray!25}{Human FATs:}\\
$\blacktriangleright$ \texttt{Sample preparation; FIB Cross Section; XEIA3+}\\
$\blacktriangleright$ \texttt{Physical Analysis; SEM; XEIA3+}\\
$\blacktriangleright$ \texttt{Others; administrative activity; REPORT GENERATION}\\
$\blacktriangleright$ \texttt{Non destructive Inspection; X-ray; XRAY DAGE XD7600NT DIAMOND FP}\\
$\blacktriangleright$ \texttt{Non destructive Inspection; SAM; SONOSCAN GEN6}\\
$\blacktriangleright$ \texttt{Non destructive Inspection; Optical microscopy; KEYENCE VHX-6000}\\
$\blacktriangleright$ \texttt{Non destructive Inspection; IR microscopy; MIC-LSCM}\\
$\blacktriangleright$ \texttt{Global fault localisation; Thermal microscopy; THEMOS1000}\\
$\blacktriangleright$ \texttt{Global fault localisation; Static Laser Techniques; PHEMOS1000}\\
$\blacktriangleright$ \texttt{Electrical Failure Verification; Continuity test; AUTOMATIC CONTINUITY TESTER-CURVE TRACER}\\
\colorbox{lightgray!25}{PT-GPT2 Medium:} \colorbox{blue!15}{$\blacktriangleright$ FATs for Grenoble}\\
$\blacktriangleright$ \texttt{\colorbox{green!10}{Sample preparation}; \colorbox{lightgray!65}{Die Delayering}; \colorbox{green!10}{XEIA3+}} \\
$\blacktriangleright$ \texttt{\colorbox{lightgray!65}{Sample preparation}; \colorbox{lightgray!65}{Die Delayering}; \colorbox{lightgray!65}{NE860}} \\
$\blacktriangleright$ \texttt{\colorbox{lightgray!65}{Sample preparation}; \colorbox{lightgray!65}{Die Delayering}; \colorbox{lightgray!65}{CHEMISTRY+}} \\
$\blacktriangleright$ \texttt{\colorbox{lightgray!65}{Sample preparation}; \colorbox{lightgray!65}{Die Delayering}; \colorbox{lightgray!65}{BINOCULAR}} \\
$\blacktriangleright$ \texttt{\colorbox{green!10}{Physical Analysis}; \colorbox{green!10}{SEM}; \colorbox{lightgray!65}{MAGELLAN}} \\
$\blacktriangleright$ \texttt{\colorbox{green!10}{Others}; \colorbox{green!10}{administrative activity}; \colorbox{green!10}{REPORT GENERATION}} \\
$\blacktriangleright$ \texttt{\colorbox{lightgray!65}{Others}; \colorbox{lightgray!65}{administrative activity}; \colorbox{lightgray!65}{REQUEST MANAGEMENT}} \\
$\blacktriangleright$ \texttt{\colorbox{green!10}{Non destructive Inspection}; \colorbox{green!10}{Optical microscopy}; \colorbox{green!10}{KEYENCE VHX-6000}} \\
$\blacktriangleright$ \texttt{\colorbox{green!10}{Global fault localisation}; \colorbox{lightgray!65}{EMMI}; \colorbox{lightgray!65}{PHEMOS1000-COMBO}} \\
$\blacktriangleright$ \texttt{\colorbox{green!10}{Global fault localisation}; \colorbox{lightgray!65}{Electrical parametrical test}; \colorbox{lightgray!65}{ELECTRICAL TOOLS}} \\
$\blacktriangleright$ \texttt{\colorbox{green!10}{Electrical Failure Verification}; \colorbox{lightgray!65}{Electrical parametrical test};\\ \colorbox{lightgray!65}{AUTOMATIC CONTINUITY TESTER-CURVE TRACER}} \\
\colorbox{lightgray!25}{PT-GPT2 Medium:} \colorbox{blue!15}{$\blacktriangleright$ FATs for Calambda}\\
$\blacktriangleright$ \texttt{\colorbox{green!10}{Sample preparation}; \colorbox{lightgray!65}{Package decap}; \colorbox{lightgray!65}{MANUAL CHEMICAL DECAP}} \\
$\blacktriangleright$ \texttt{\colorbox{lightgray!65}{Sample preparation}; \colorbox{lightgray!65}{Package decap}; \colorbox{lightgray!65}{B800\_FA-LD01}} \\
$\blacktriangleright$ \texttt{\colorbox{green!10}{Non destructive Inspection}; \colorbox{green!10}{X-ray}; \colorbox{lightgray!65}{B800\_FA-XR01}} \\
$\blacktriangleright$ \texttt{\colorbox{green!10}{Physical Analysis}; \colorbox{lightgray!65}{Optical inspection}; \colorbox{lightgray!65}{OLYMPUS MICROSCOPE SZ40 (ROS005)}} \\
$\blacktriangleright$ \texttt{\colorbox{green!10}{Electrical Failure Verification}; \colorbox{green!10}{Continuity test}; \colorbox{lightgray!65}{TEKTRONIX 370A CURVE TRACER}} \\
$\blacktriangleright$ \texttt{\colorbox{green!10}{Non destructive Inspection}; \colorbox{green!10}{Optical microscopy}; \colorbox{lightgray!65}{OLYMPUS MICROSCOPE SZ40 (ROS005)}} \\
$\blacktriangleright$ \texttt{\colorbox{green!10}{Electrical Failure Verification}; \colorbox{green!10}{Continuity test}; \colorbox{lightgray!65}{KEITHLEY PARAMETRIC ANALYZER}} \\
\colorbox{lightgray!25}{PT-GPT2 Medium:} \colorbox{blue!15}{$\blacktriangleright$ FATs for Shenzhen}\\
$\blacktriangleright$ \texttt{\colorbox{green!10}{Sample preparation}; \colorbox{lightgray!65}{Package decap}; \colorbox{lightgray!65}{DE-CAPPING CHEMICALDECAP ACID}} \\
$\blacktriangleright$ \texttt{\colorbox{lightgray!65}{Sample preparation}; \colorbox{lightgray!65}{Die extraction wire and glue removal};\\ \colorbox{lightgray!65}{DE-CAPPING CHEMICALDECAP ACID}} \\
$\blacktriangleright$ \texttt{\colorbox{green!10}{Non destructive Inspection}; \colorbox{green!10}{X-ray}; \colorbox{lightgray!65}{3D X-RAY}} \\
$\blacktriangleright$ \texttt{\colorbox{green!10}{Non destructive Inspection}; \colorbox{green!10}{Optical microscopy}; \colorbox{lightgray!65}{VHX 5000}} \\
$\blacktriangleright$ \texttt{\colorbox{green!10}{Electrical Failure Verification}; \colorbox{green!10}{Continuity test}; \colorbox{lightgray!65}{CURVE TRACER 2}}
\normalsize

The triplets generated for the different locations differ by the type of equipment used for failure analysis and only slightly in step and substep technique. It implies that specifying the location of FA in the failure description, consequently improves the accuracy of both SS-FATG and LS-FATG. 

\subsection{Discussion}
\begin{figure}[!ht]
			\centering
			\subfloat[\centering \small LESE-3 cost matrix before flipping equipment. \texttt{LESE-3} $= 100.00\%$.]{{\includegraphics[width = 4.40cm]{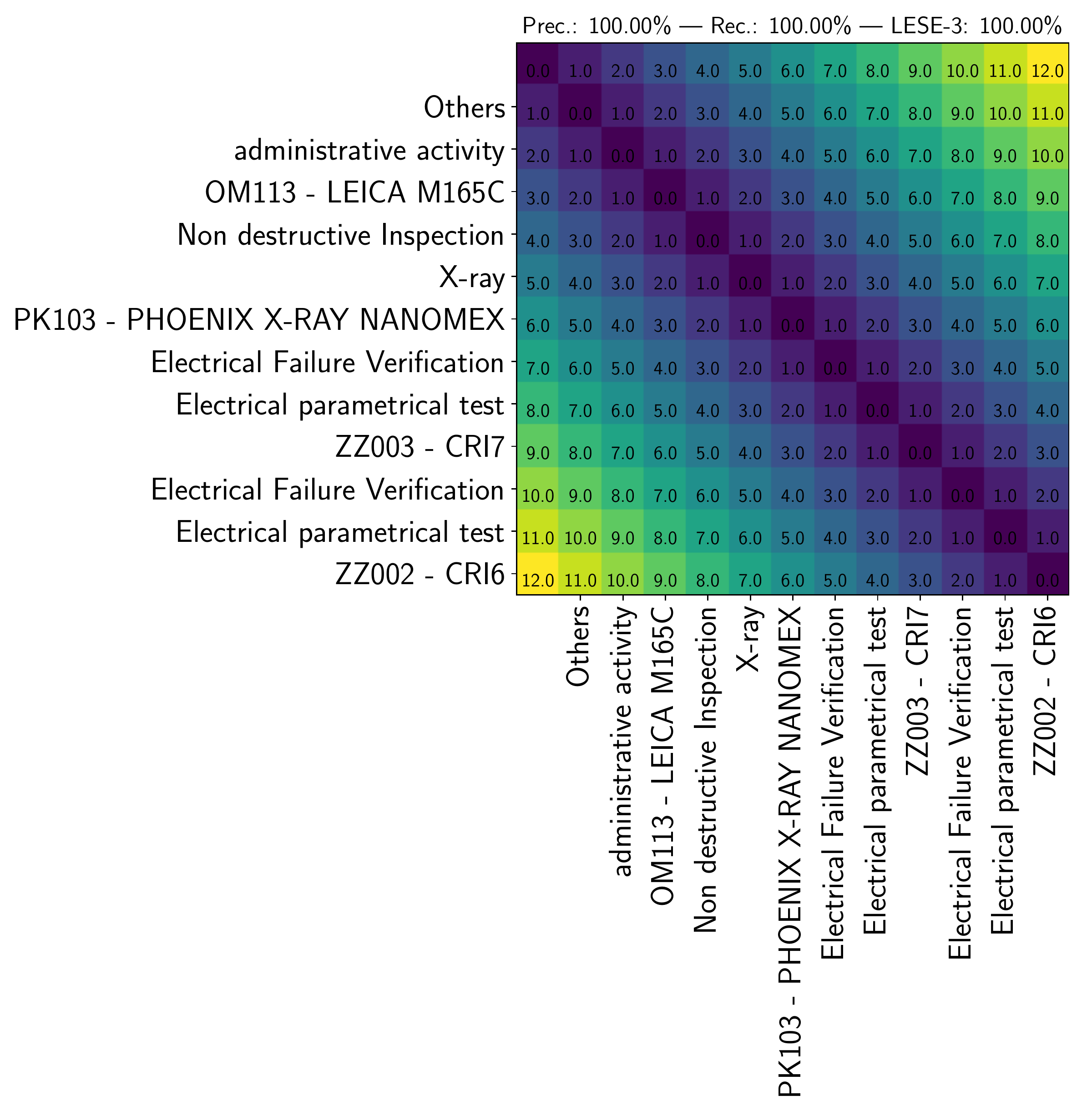}}}\hspace{.5cm}
			\subfloat[\centering \small LESE-1 cost matrix after flipping equipment. \texttt{LESE-1} $= 83.33\%$.]{{\includegraphics[width = 4.37cm]{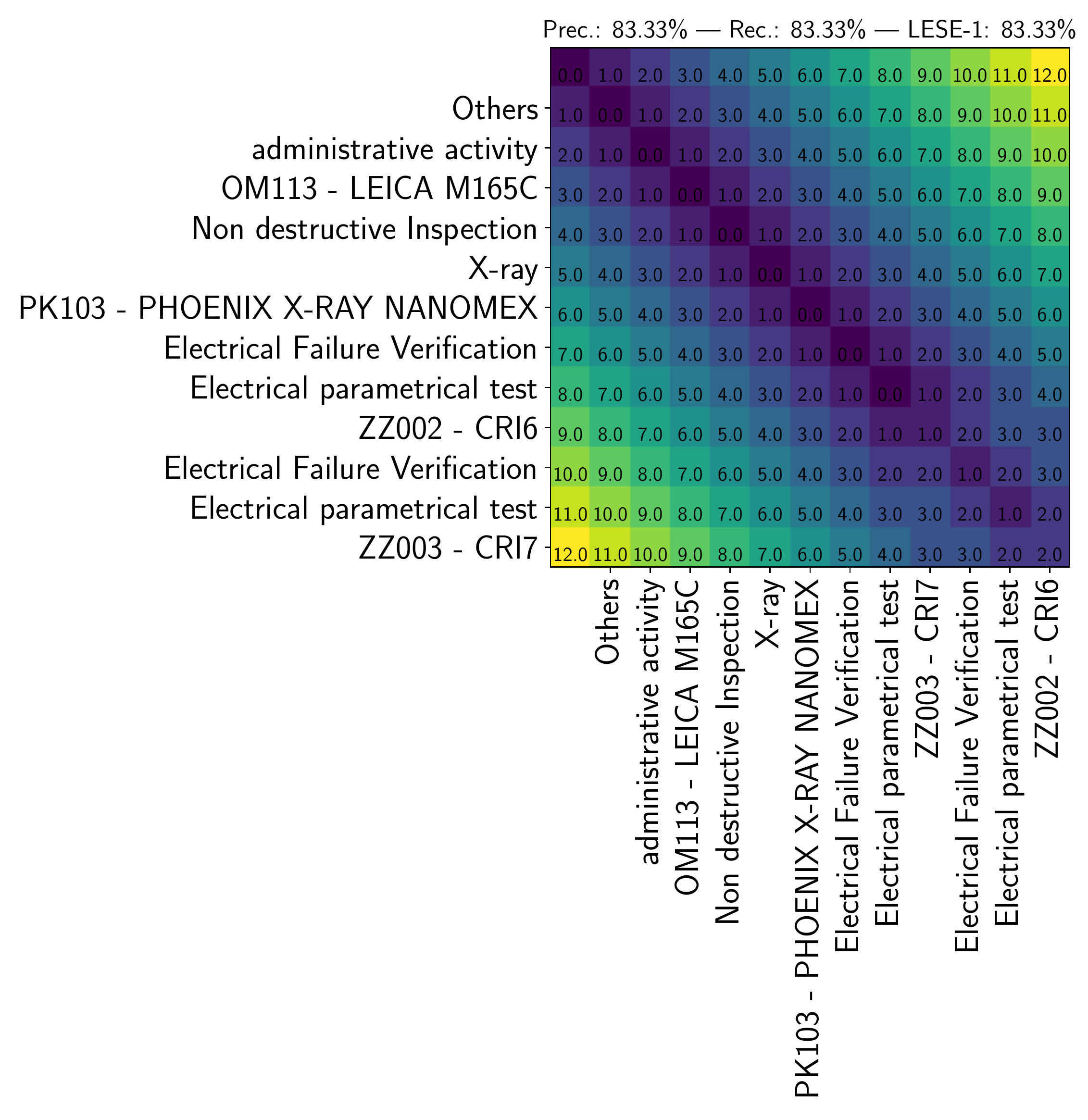}}} \hspace{.5cm}
			\subfloat[\centering \small LESE-3 cost matrix after flipping equipment. \texttt{LESE-3} $= 50.00\%$.]{{\includegraphics[width = 4.45cm]{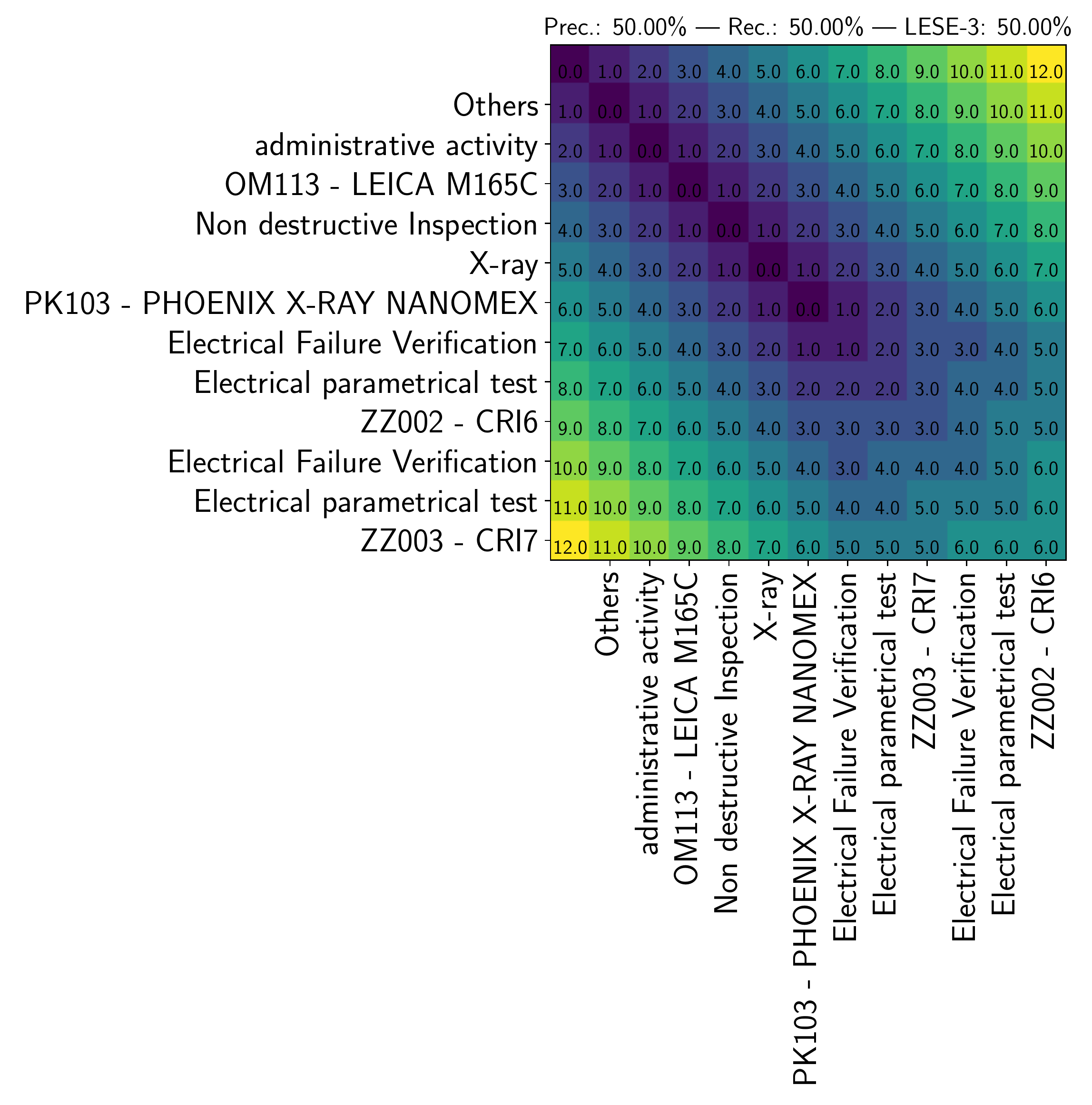}}}
			\caption{\label{lese_1}\small LESE-3 cost matrix before and after flipping equipment in triplet generated by GPT2. observe the final Levenshstein distance before flipping is $2$ and $6$ after flipping. The LESE-3 (F1-score) reflects this change in equipment and correlates with human evaluation.}
\end{figure}

We examine the near human accuracy of LESE score over other evaluation metric. The generation results from Figure (\ref{lese_1}a) indicates a perfect LESE-$1$ and LESE-$3$ score of $100\%$; BLEU-1 and BLEU-$3$ scores of $100\%$ and ROUGE-$1$ and ROUGE-L (F1-score) of $99.9\%$. However, if we flip the equipment in the last two triplets (Figure (\ref{lese_1}b \& \ref{lese_1}c)), the LESE-1 and LESE-3 scores become ($83.33\%$, $50.00\%$); BLEU-1 and BLEU-$3$ scores ($100\%$, $90.9\%$); ROUGE-$1$ and ROUGE-L (F1-score) are ($88.88\%$, $88.88\%$). We discover that both BLEU and ROUGE scores suffer significant setbacks due to this small flip and do not accurately quantify the change. Furthermore, BLEU-1 does not seem to distinguish the difference between equipment in the last two FATs, since it only seeks to find the intersection between sets in hypothesis and reference. 
LESE-1 and LESE-3 accurately measures the difference in this flip to a human level. Observe how the Levenshstein distance changes the heat on the images along the diagonal from index $9$ in Figure (\ref{lese_1}b) and index $7$ on Figure (\ref{lese_1}c). This changes indicate the time of occurrence of the $n$-gram Levenshstein deletion, insertion or substitution operations.
\begin{figure}[!ht]
\centering
		\subfloat[\centering \small LESE-3 cost matrix before transposing triplets. \texttt{LESE-3} $=$ \texttt{LESE-1} $= 40.00\%$, \texttt{Precision} $= 25.00\%$ \& \texttt{Recall} $= 100.00\%$.]{{\includegraphics[width = 8.60cm]{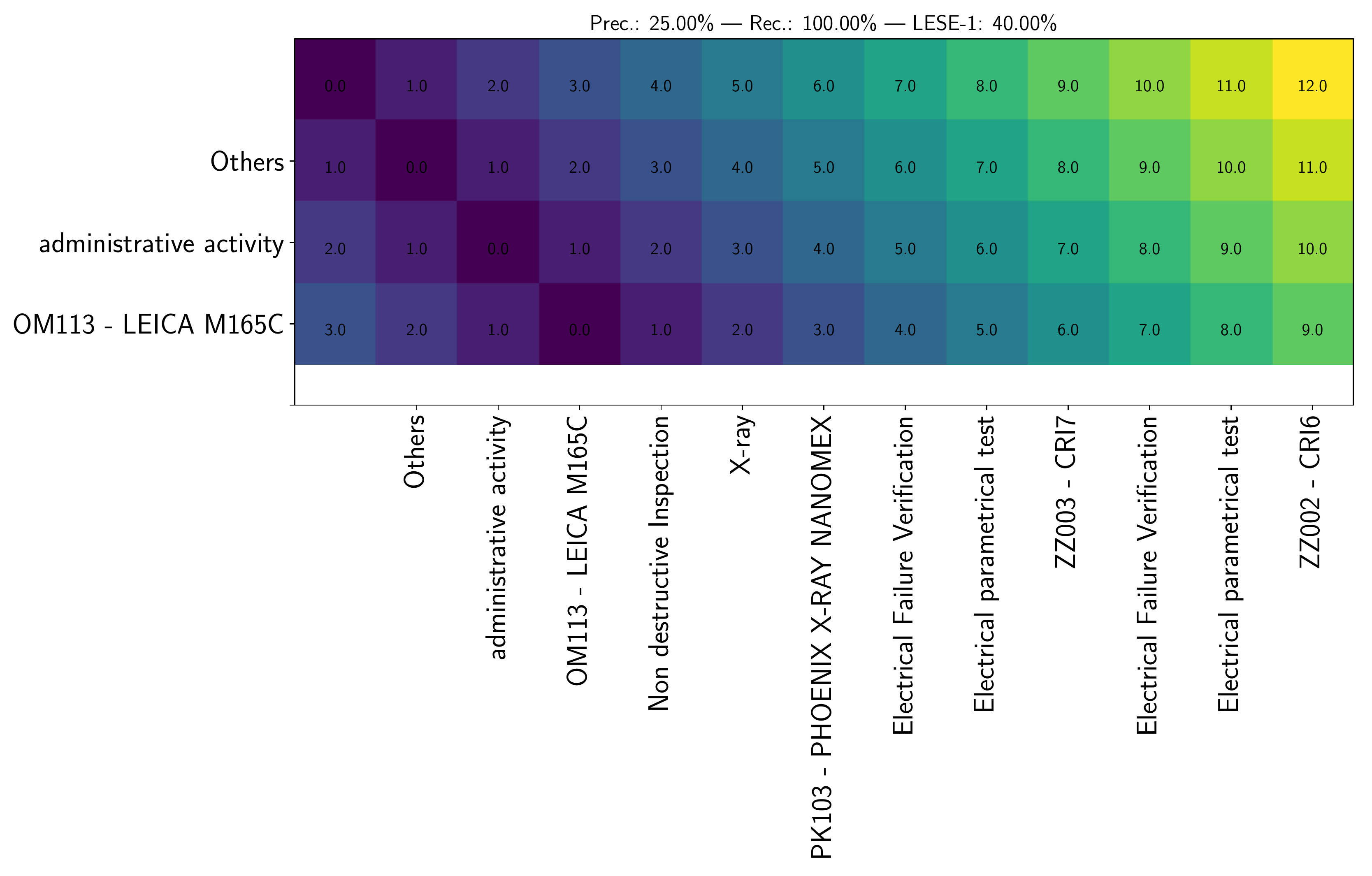}}} \hspace{0.3cm}
		\subfloat[\centering \small LESE-3 cost matrix after transposing triplets. \texttt{LESE-3} $=$ \texttt{LESE-1} $= 40.00\%$, \texttt{Precision} $= 100.00\%$ \& \texttt{Recall} $= 25.00\%$.]{{\includegraphics[width = 5.60cm]{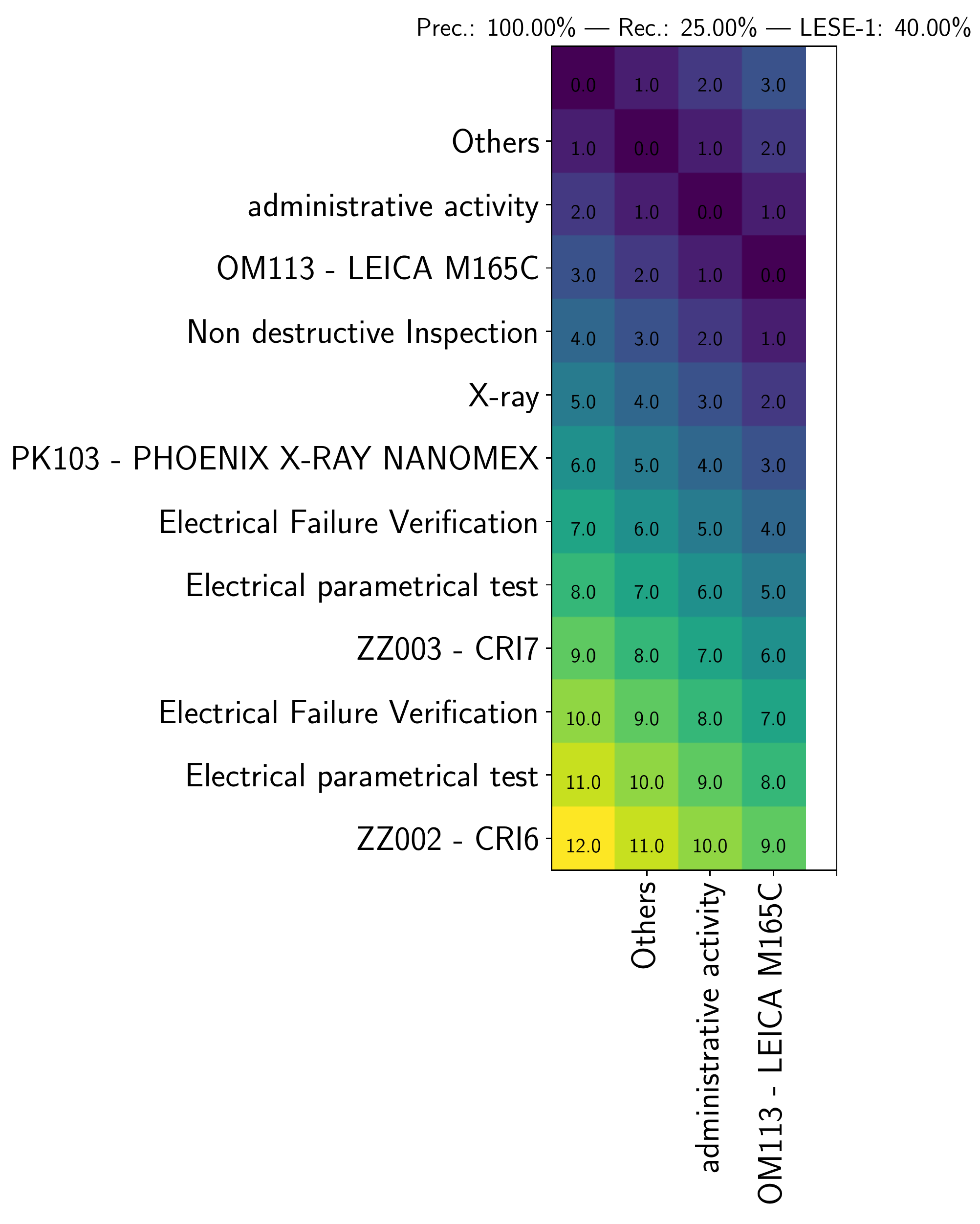}}}
		\caption{\label{lese_2}\small LESE-3 cost matrix with unequal $3$-gram length of reference and hypothesis.}
\end{figure}
\begin{figure}[!ht]
\centering
		\subfloat[\centering \small LESE-3 cost matrix. \texttt{LESE-3}  $= 40.00\%$, \texttt{Precision} $= 25.00\%$ \& \texttt{Recall} $= 100.00\%$.]{{\includegraphics[width = 6.83cm]{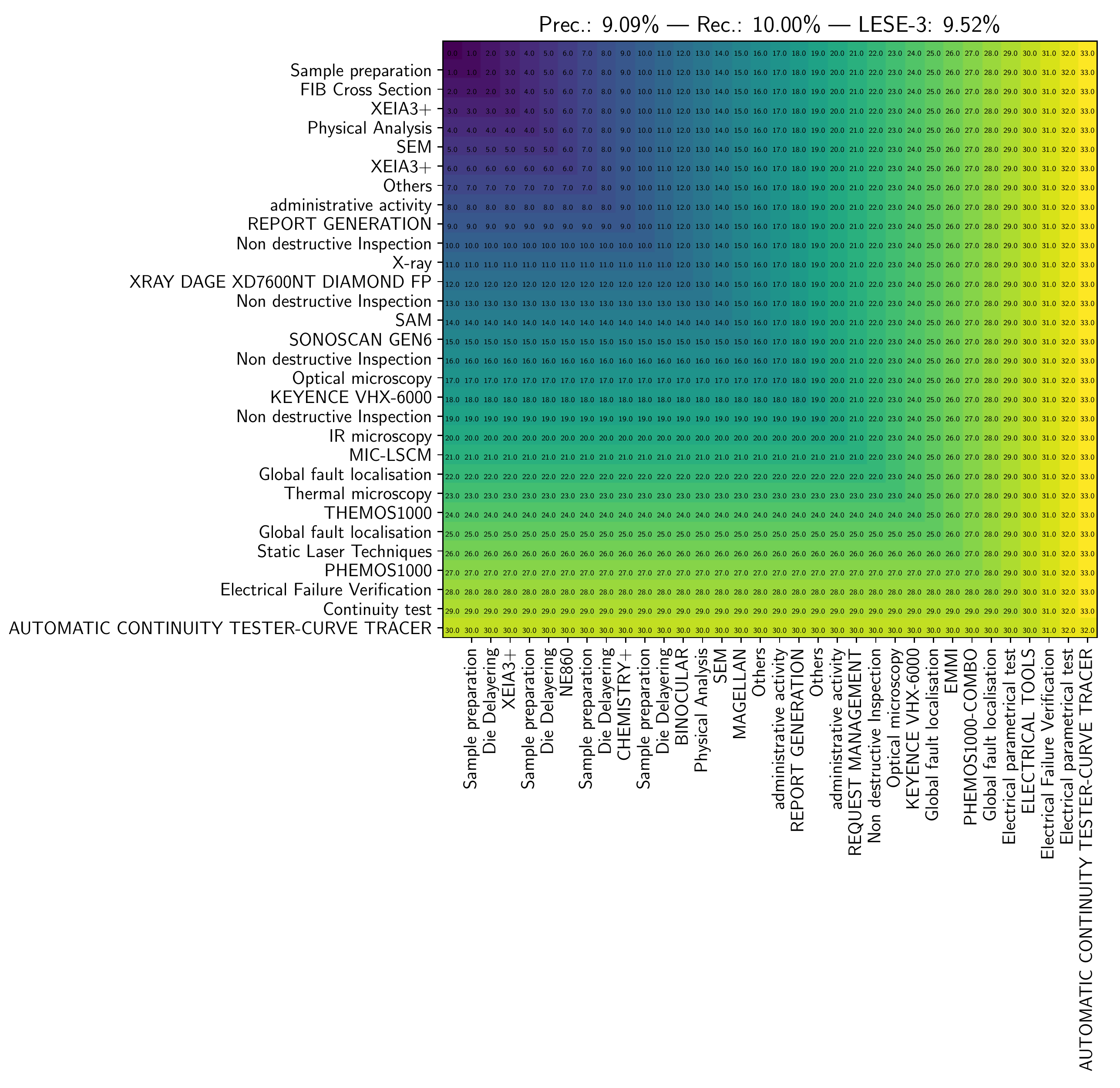}}} \hspace{0.3cm}
		\subfloat[\centering \small LESE-3 cost matrix. \texttt{LESE-3} $= 40.00\%$, \texttt{Precision} $= 100.00\%$ \& \texttt{Recall} $= 25.00\%$.]{{\includegraphics[width = 6.60cm]{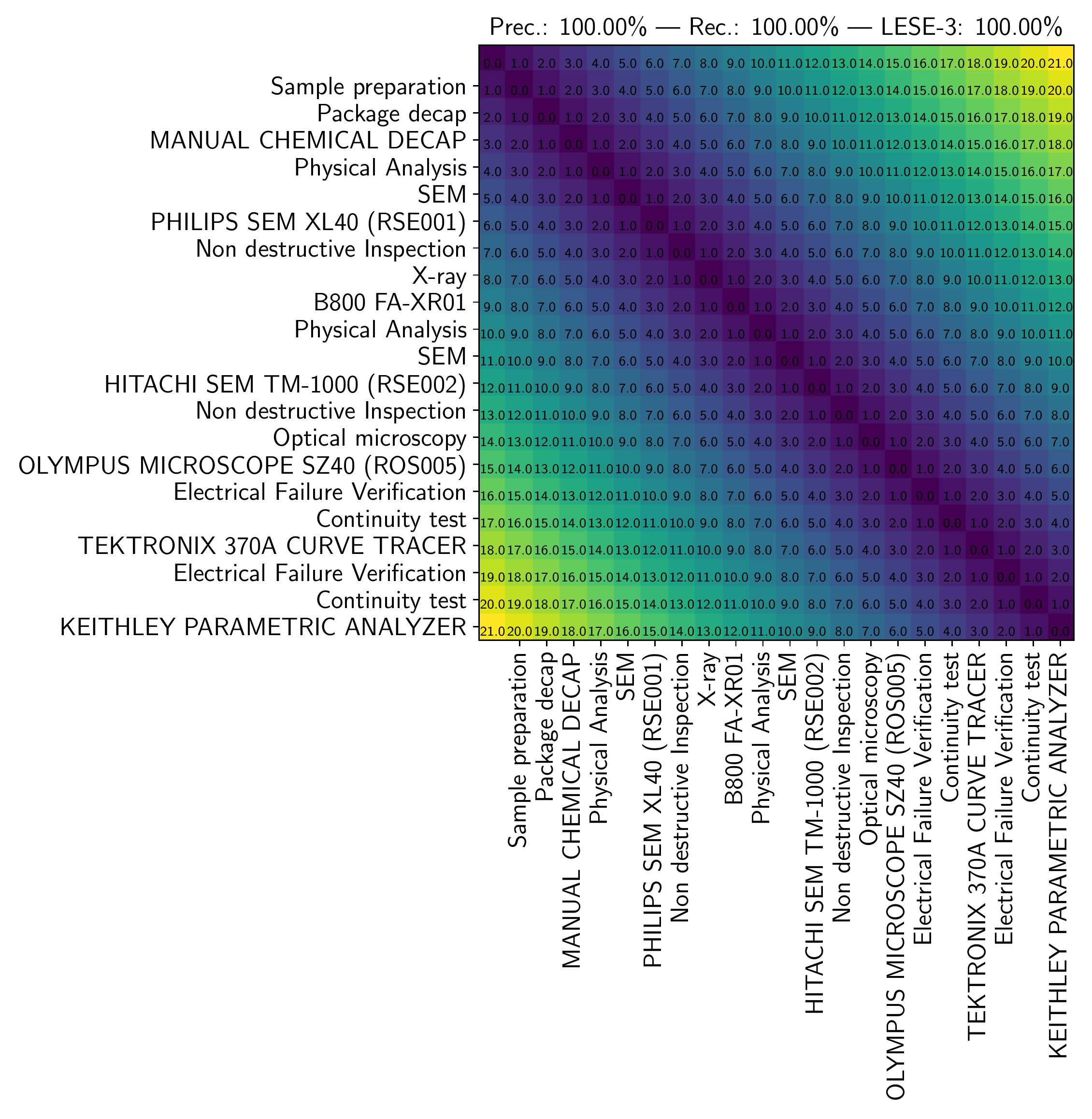}}}
		\caption{\label{lese_3}\small LESE-3 cost matrix with unequal $3$-gram length of reference and hypothesis.}
\end{figure}

When \texttt{|r|}$_\texttt{n-gram} = 1$ and less than \texttt{|h|}$_\texttt{n-gram}$ or \texttt{|h|}$_\texttt{n-gram} = 1$ and less than \texttt{|r|}$_\texttt{n-gram}$, the LESE-1 score is computed (See Figure (\ref{lese_2})). A python implementation of the LESE-N algorithm is presented in the appendix. If we transpose Figure (\ref{lese_2}a) to (\ref{lese_2}b), the LESE-3 score is unchanged, however, precision becomes $100\%$ while recall is $25\%$. Other examples of LESE-3 evaluation are presented in Figures (\ref{lese_3}).

\section{Conclusion}
We evaluate the efficiency of pre-trained transformer models for the downstream task of failure analysis triplet generation. We observe that forward only auto-regressive modelling used in GPT2 and GPT3 gives it excellent capabilities to generate structured data. When adapted for FATG, GPT2 ROUGE scores outperforms other benchmarks for generating both short and long FATs, $\lambda$'s. Since, ROUGE, BLEU and METEOR scores do not accurately convey human-expert evaluation, we introduce Levenshstein Sequential Evaluation (LESE) metric. LESE-N performs on par with human expert judgment for different test cases. GPT2-Medium and Large models perform best on LESE-1 and LESE-3 triplet scores, which corresponds to human judgement.

Fine-tuned BART generates very short triplets and seeks contextual representation of triplets making it unfit for structured long-text sequences while GPT3, generates long story-telling-like data that do not necessarily follow known expert failure analysis FATs. 
\bibliographystyle{apalike}
\bibliography{sample}
\appendix
\appendix
\onecolumn
\subsection{Abbreviations}
\begin{table}[!ht]
    \centering
    \centering
    \setlength{\tabcolsep}{0.5em} 
    {\renewcommand{\arraystretch}{1.7}
    \scriptsize
    \begin{tabular}{l|l}
    \Xhline{1.5pt}
        \cellcolor[HTML]{EFEFEF} Abbreviation & \cellcolor[HTML]{EFEFEF} Meaning \\ \hline
         RCA & Root Cause Analysis \\ \hline
         FMEA & Failure Mode and Effects Analysis \\ \hline
         RE & Reliability Engineering \\ \hline
         FA & Failure Analysis \\ \hline
         FRACAS & Failure Reporting Analysis and Corrective Action System \\ \hline
         FDR & Failure Description \\ \hline
         FATs & Failure Analysis Triplets \\ \hline
         FATG & Failure Analysis Triplet Generation \\ \hline
         LM & Language Model \\ \hline
         PLMs & Pre-trained Language Models \\ \hline
         Seq2Seq & Sequence-to-Sequence \\ \hline
         RNN & Recurrent Neural Network \\ \hline
         LSTM & Long Short Term Memory \\ \hline
         GRNN & Gated Recurrent Neural Network \\ \hline
         PFFN & Position-wise Feed-Forward Network \\ \hline
         PE & Position Embedding \\ \hline
         BART & Bidirectional Auto-Regressive Transformer \\ \hline
         GPT & Generative Pre-trained Transformers \\ \hline
         BLEU & Bilingual Evaluation Understudy \\ \hline
         ROUGE & Recall-Oriented Understudy for Gisting Evaluation \\ \hline
         LESE & Levenshstein Sequential Evaluation \\  \hline
         SS-FATG & Short sequence FATG \\ \hline
         LS-FATG & Long sequence FATG \\ \Xhline{1.5pt}
    \end{tabular}
    \caption{A list of abbreviations and their respective meaning}
    \label{tab:my_label}
    }
\end{table}
\clearpage
\subsection{Python Implementation of LESE-N Algorithm.} \label{less_algo}

\begin{lstlisting}[language=Python, caption=Python Implementation of LESE-N Algorithm.]
import numpy as np
class LESE:
    def __init__(self, a, b, n_gram):
        '''LESE: LEvenshstein Sequential Evaluation metric

        Parameters
        ----------
        a : str or list
            Reference string sequence or list of string sequence.
        b : str or list
            Hypothesis string sequence or list of string sequence.
        n_gram : int, optional
            n-gram size. The default is 3. See lese(..) sub module
            
        Returns
        -------
        None
        '''
        self.a = a
        self.b = b
        self.n_gram = n_gram
        self.lese(self.a, self.b, self.n_gram)
    
    def precision_lev_(self, D_i_j, n_a, n_b, n):
        max_n_a_n_b = max(n_a, n_b)
        lev_seq = D_i_j//n
        prec = max(0, abs((max_n_a_n_b - lev_seq))/n_b)
        return prec
    
    def recall_lev_(self, D_i_j, n_a, n_b, n):
        max_n_a_n_b = max(n_a, n_b)
        lev_seq = D_i_j//n 
        rec = max(0, abs((max_n_a_n_b - lev_seq))/n_a)
        return rec
    
    def f_score_lev_(self, D_i_j, n_a, n_b, n, beta = 1.):
        '''Sequential Levenshstein F1- score
    
        Parameters
        ----------
        D_i_j : float
            Levenshstein distance.
        n_a : int
            N-gram length of reference sequence.
        n_b : int
            N-gram length of hypothesis sequence.
        n : int
            N-gram size.
        beta : float, optional
            beta weight for balancing precision-recall. The default is 1.
    
        Returns
        -------
        float
            Precision.
        float
            Recall.
        float
            f_score.
        '''
        precision = self.precision_lev_(D_i_j, n_a, n_b, n)
        recall = self.recall_lev_(D_i_j, n_a, n_b, n)
        if precision == 0. and recall == 0.:
            return 0., 0., 0.
        else:
            f_score = ((1+ np.square(beta))*precision*recall)/(np.square(beta)*precision+recall)
        return precision, recall, f_score
    
    def lese(self, a, b, n = 3):
        '''LESE: LEvenshstein Sequential Evaluation metric main module
    
        Parameters
        ----------
        a : str or list
            Reference string sequence or list of string sequence.
        b : str or list
            Hypothesis string sequence or list of string sequence.
        n : int, optional
            n-gram size. The default is 3.
    
        Returns
        -------
        Tuple
            Tuple(Levenshstein distance, (precision, recall, f_1_score)).
            
        Complexity
        ----------
        Time: O(M*N)
        Space: O(M*N)
    
        '''
        if isinstance(a, str) and isinstance(b, str):
            n_a = len(a)
            n_b = len(b)
            if(len(a)//n) ==1: n = 1
            else: pass
            if(len(b)//n) ==1: n = 1
            else: pass
            n_gram_a = len(a)//n #n-gram in reference
            n_gram_b = len(b)//n #n-gram in hypothesis
            if n_b == 0:
                self.levenshstein_distance = n_a
                self.precision_, self.recall_, self.f_score_ = (0.0, 0.0, 0.0)
                return self.levenshstein_distance, self.precision_, self.recall_, self.f_score_
            elif n_a == 0:
                self.levenshstein_distance = n_b
                self.precision_, self.recall_, self.f_score_ = (0.0, 0.0, 0.0)
                return self.levenshstein_distance, self.precision_, self.recall_, self.f_score_
            else:
                pass
        elif isinstance(a, list) and isinstance(b, list):
            a = [x for x in a if x != 'nan' if x != ' ' if x != '']
            b = [x for x in b if x != 'nan' if x != ' ' if x != '']
            n_a = len(a)
            n_b = len(b)
            if(len(a)//n) ==1: n = 1
            else: pass
            if(len(b)//n) ==1: n = 1
            else: pass
            n_gram_a = len(a)//n
            n_gram_b = len(b)//n
            if n_b == 0:
                self.levenshstein_distance = n_a
                self.precision_, self.recall_, self.f_score_ = (0.0, 0.0, 0.0)
                return self.levenshstein_distance, self.precision_, self.recall_, self.f_score_
            elif n_a == 0:
                self.levenshstein_distance = n_b
                self.precision_, self.recall_, self.f_score_ = (0.0, 0.0, 0.0)
                return self.levenshstein_distance, self.precision_, self.recall_, self.f_score_
            else:
                pass
        
        self.D = np.zeros((n_a + 1, n_b + 1))
        # Initialising first row using the length of string/list a: reference
        for i in range(n_a + 1):
            self.D[i][0] = i
        # Initialising first column using the length of string/list b: hypothesis
        for j in range(n_b + 1):
            self.D[0][j] = j
        #DP version
        for i in range(1, n_a + 1):
            for j in range(1, n_b + 1):
                if a[i-1:i+n-1] == b[j-1:j+n-1]:
                    self.D[i][j] = self.D[i - 1][j - 1]
                else:
                    insertion = 1 + self.D[i][j - 1] #insertion operation
                    deletion = 1 + self.D[i - 1][j] #deletion operation
                    substitution = 1 + self.D[i - 1][j - 1] #substitution operation
                    #select optimal cost, optimal cost == minimum cost of operation
                    self.D[i][j] = min(insertion, deletion, substitution)
        self.levenshstein_distance = self.D[i][j]
        self.precision_, self.recall_, self.f_score_ = self.f_score_lev_(self.D[i][j], n_gram_a, n_gram_b, n)
        return self.levenshstein_distance, self.precision_, self.recall_, self.f_score_
\end{lstlisting}

\end{document}